\documentclass[conference]{IEEEtran}
\usepackage{cite}
\usepackage{amsmath,amssymb,amsfonts}
\usepackage{graphicx}
\usepackage{textcomp}
\usepackage{xcolor}
\usepackage{fancyhdr}
\usepackage[hyphens]{url}

\def\BibTeX{{\rm B\kern-.05em{\sc i\kern-.025em b}\kern-.08em
    T\kern-.1667em\lower.7ex\hbox{E}\kern-.125emX}}

\usepackage{hyperref}
\usepackage{color}
\usepackage{soul}
\usepackage{mathptmx} 
\usepackage{fancyhdr}
\usepackage[normalem]{ulem}
\usepackage[hyphens]{url}
\usepackage{flushend}
\usepackage{caption}
\usepackage{subfig}
\usepackage{blindtext}
\usepackage{algorithm}
\usepackage{algorithmic}
\usepackage{booktabs}
\usepackage{amsmath}
\usepackage{amsfonts}
\usepackage{amssymb}
\usepackage{mathtools}
\usepackage{makecell}
\usepackage{color}
\usepackage{multirow}
\usepackage[flushleft]{threeparttable}
\usepackage{float}
\usepackage{enumitem}
\usepackage{bm}
\usepackage{comment}
\usepackage{tablefootnote}
\usepackage{soul}
\usepackage{textgreek}
\usepackage[noabbrev]{cleveref}

\newcommand{\ra}[1]{\renewcommand{\arraystretch}{#1}}

\usepackage{pifont}
\newcommand\mynuma[1]{\ifcase#1 \or \ding{172}\or \ding{173}\or
  \ding{174}\or \ding{175}\or \ding{176}\or \ding{177}%
  \or \ding{178}\or \ding{179}\or \ding{180}\or \ding{181}\else *\fi\relax}
\newcommand\mynumb[1]{\ifcase#1 \or \ding{182}\or \ding{183}\or
  \ding{184}\or \ding{185}\or \ding{186}\or \ding{187}%
  \or \ding{188}\or \ding{189}\or \ding{190}\or \ding{191}\else *\fi\relax}

\newcommand{\PLH}{{\mkern-2mu\times\mkern-2mu}}

\makeatletter

\makeatletter

\IEEEoverridecommandlockouts

\title{SmartExchange: Trading Higher-cost Memory Storage/Access for Lower-cost Computation
\thanks{$\ast$ denotes equal contribution.}
}
\author{
\IEEEauthorblockN{Yang Zhao$^{1 \ast}$, Xiaohan Chen$^{2 \ast}$, Yue Wang$^1$, Chaojian Li$^1$, Haoran You$^1$, Yonggan Fu$^1$,\\
Yuan Xie$^3$, Zhangyang Wang$^2$, Yingyan Lin$^1$}
\IEEEauthorblockA{$^1$\textit{Department of Electrical and Computer Engineering, Rice University}}
\IEEEauthorblockA{$^2$\textit{Department of Computer Science and Engineering, Texas A\&M University}}
\IEEEauthorblockA{$^3$\textit{Department of Electrical and Computer Engineering, UC Santa Barbara}}
\IEEEauthorblockA{\textit{Email: \{zy34, yw68, cl114, hy34, yf22,  yingyan.lin\}@rice.edu, \{chernxh, atlaswang\}@tamu.edu, \{yuanxie\}@ucsb.edu}}
}

\begin{document}
\maketitle


\begin{abstract}
We present \textit{SmartExchange}, an algorithm-hardware co-design framework to trade higher-cost memory storage/access for lower-cost computation, for energy-efficient inference of deep neural networks (DNNs). We develop a novel algorithm to enforce a specially favorable DNN weight structure, where each layerwise weight matrix can be stored as the product of a small basis matrix and a large sparse coefficient matrix whose non-zero elements are all power-of-2. To our best knowledge, this algorithm is the first formulation that integrates three mainstream model compression ideas: \textit{sparsification or pruning}, \textit{decomposition}, and \textit{quantization}, into one unified framework. The resulting sparse and readily-quantized DNN thus enjoys greatly reduced energy consumption in data movement as well as weight storage. 
On top of that, we further design a dedicated accelerator to fully utilize the \textit{SmartExchange}-enforced weights to improve both energy efficiency and latency performance. Extensive experiments show that 1) \ul{on the algorithm level}, \textit{SmartExchange} outperforms state-of-the-art compression techniques, including merely \textit{sparsification or pruning}, \textit{decomposition}, and \textit{quantization}, in various ablation studies based on nine models and four datasets; and 2) \ul{on the hardware level}, \textit{SmartExchange} can boost the energy efficiency by up to 6.7$\times$ and reduce the latency by up to 19.2$\times$ 
over four state-of-the-art DNN accelerators, when benchmarked on seven DNN models (including four standard DNNs, two compact DNN models, and one segmentation model) and three datasets. 
\end{abstract}



 \vspace{-1em}
\section{Introduction}\label{sec:introduction}
 \vspace{-0.3em}
We have recently witnessed the record-breaking performance of deep neural networks (DNNs) together with a tremendously growing demand to bring DNN-powered intelligence into resource-constrained edge devices \cite{liu2018adadeep,wang2019dual}, which have limited energy and storage resources. However, as the excellent performance of modern DNNs comes at a cost of a huge number of parameters which need external dynamic random-access memory (DRAM) for storage, the prohibitive energy consumed by the massive data transfer between DRAM and on-chip memories or processing elements (PEs) makes DNN deployment non-trivial. The resource-constrained scenarios in edge devices motivate more efficient domain-specific accelerators for DNN inference tasks \cite{chen2014diannao,RDSEC,Ziyun-ISSC2019,chen2017eyeriss, Boris_ISSCC}.

The DNN accelerator design faces one key challenge: \textit{how to alleviate the heavy data movement}? Since DNN inference mainly comprises multiply-and-accumulate (MAC) operations, it has little data dependency and can achieve high processing throughput via parallelism. However, these MAC operations incur a significant amount of data movement, due to read/write data accesses, which consumes considerable energy and time, and sometimes surprisingly significant (especially when the inference batch size is small or just one). Take DianNao as an example, more than 95\% of the inference energy is consumed by data movements associated with the DRAM~\cite{chen2017eyeriss,chen2014diannao,Ziyun-JSSC2019}. Therefore, minimizing data movements is the key to improve the energy/time efficiency of DNN accelerators.


\begin{figure}[!t]
    \centerline{\includegraphics[width=78mm]{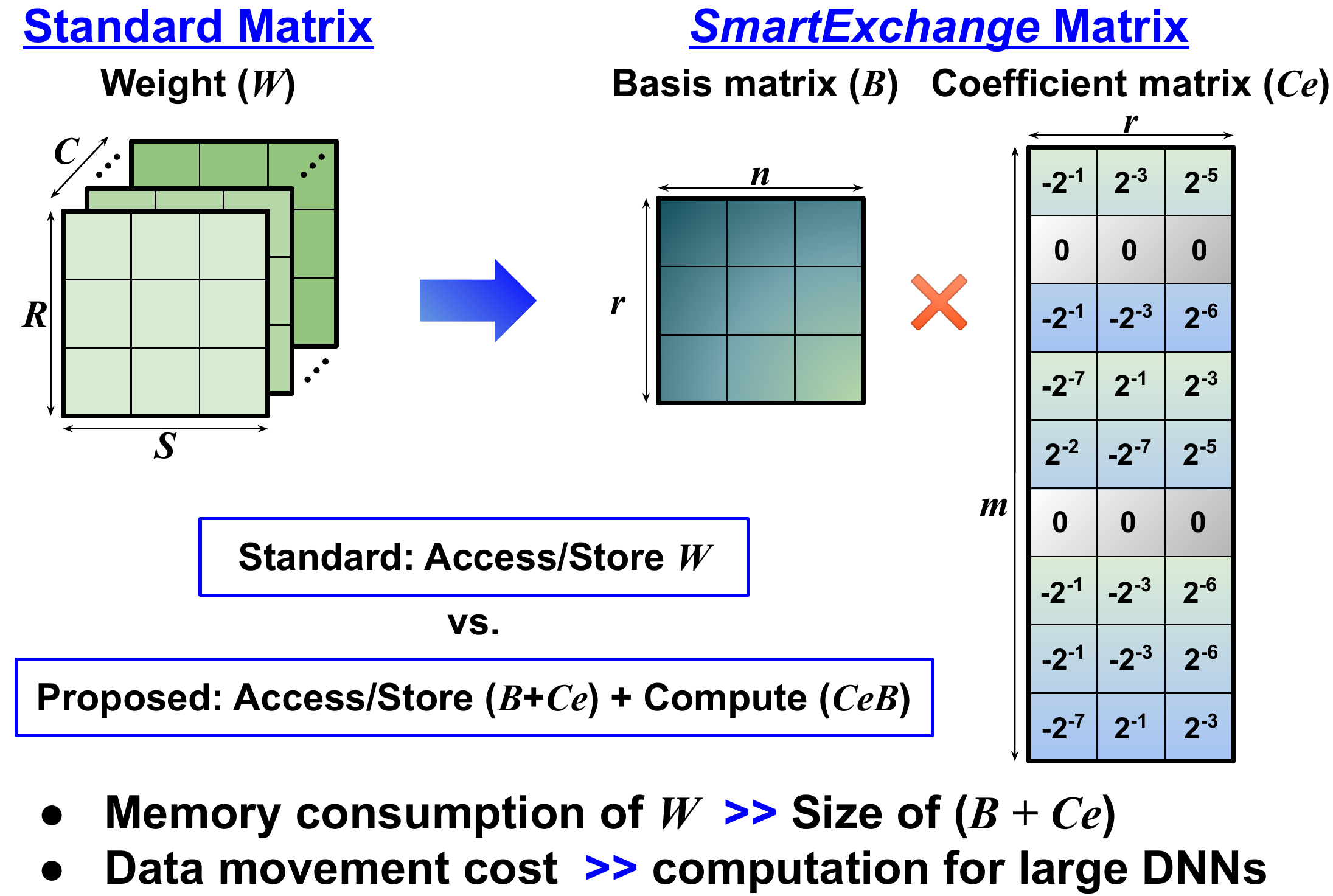}}
    \vspace{-0.5em}
    \caption{The proposed \textit{SmartExchange}'s weight representation.}
    \label{fig:SE_concept}
    \vspace{-1.8em}
\end{figure}

To address the aforementioned challenges, we propose the \textit{SmartExchange} solution in the spirit of algorithm-hardware co-design that strives to trade higher-cost memory storage/access for lower-cost computation to largely avoid the dominant data movement cost in DNN accelerators. In this particular work, we present a novel \textit{SmartExchange}
algorithm for aggressively reducing both 1) the energy consumption of data movement and 2) storage size associated with DNN weights, both of which are major limiting factors when deploying DNN accelerators into resource-constrained devices.  

Our solution represents a layer-wise DNN weight matrix as the product of a small basis matrix and a large coefficient matrix as shown in Figure~\ref{fig:SE_concept}.
We then simultaneously enforce two strong structural properties on the coefficient matrix: 1) \textit{sparse}: most elements are zeros and 2) \textit{readily-quantized}: the non-zero elements take only power-of-2 values, which have compact bit representations and turn the multiplications in MAC operations into much lower-cost shift-and-add operations. We then develop an efficient \textit{SmartExchange} algorithm 
 blended with a re-training process. Experiments using nine models on four datasets  indicate that such favorable decomposed and compact weight structures can be achieved using our proposed algorithm.
    
To fully leverage \textit{SmartExchange} algorithm's potential, we further develop a dedicated DNN accelerator that takes advantage of the much reduced weight storage and readily-quantized weights resulting from the algorithm to enhance hardware acceleration performance. Experiments show that the proposed accelerator outperforms state-of-the-art DNN accelerators in terms of acceleration energy efficiency and latency by up to 6.7$\times$ and 19.2$\times$, respectively. 
Our contributions are summarized as three-fold: 
\begin{itemize}

\item \textbf{Our overall innovation} is an algorithm-hardware co-design framework harmonizing algorithm and hardware level innovations for maximizing the acceleration performance and task accuracy. Specifically, we first identify opportunities for saving processing energy and time in the hardware level, including reducing DRAM accesses and taking advantage of structured weight and activation sparsity, and then enforce corresponding favorable patterns/structures in the algorithm level together with dedicated efforts in the accelerator architecture to aggressively improve acceleration energy efficiency and latency. 

\item \textbf{Our algorithm-level contribution} is a \textit{SmartExchange} 
algorithm that is designed with strong hardware awareness. It for the first time unifies the ideas of weight pruning, weight decomposition, and quantization, leading to a highly compact weight structure that boosts acceleration speed and energy efficiency at inference with a $\le$2\% accuracy loss. Equipped with re-training, the effectiveness of the \textit{SmartExchange} algorithm is benchmarked on large datasets and state-of-the-art DNNs. 


\item \textbf{Our hardware-level contribution} is a dedicated accelerator designed to fully utilize the \textit{SmartExchange} algorithm-compressed \& quantized DNNs to minimize both inference energy and latency. We verify and optimize this accelerator using dedicated simulators validated against RTL designs. Experiments show that the proposed accelerator achieves up to $6.7\times$ better energy efficiency and $19.2\times$ speedup, over state-of-the-art designs.
\end{itemize}

The rest of the paper is organized as follows. Section II introduces the background and motivation. Section III describes the problem formulation and the \textit{SmartExchange} algorithm. Section IV presents the dedicated accelerator that aims to amplify the algorithmic gains. Section V shows extensive experiments to manifest the benefits of both the algorithm and the accelerator of \textit{SmartExchange}. Sections VI and VII summarize related works and  our conclusion, respectively.


 \vspace{-0.8em}
\section{Background and Motivation}\label{sec:Background}
 \vspace{-0.5em}
\subsection{Basics of Deep Neural Networks}
Modern DNNs usually consist of a cascade of multiple convolutional (CONV), pooling, and fully-connected (FC) layers through which the inputs are progressively processed. The CONV and FC layers can be described as:
\begin{equation}
    \begin{split}
    &\textit{\textbf{O}}[c_o][e][f]=\\
    & \sigma(\sum_{c_i=0}^{C-1}{\sum_{k_r=0}^{R-1}{\sum_{k_s=0}^{S-1}{\textit{\textbf{W}}[c_o][c_i][k_r][k_s]}\times \textit{\textbf{I}}[c_i][eU+k_r][fU+k_s]}}+\textbf{B}[c_i])\\
    & 0\le c_o < M,~0\le e < E, 0\le f < F, \nonumber
    \end{split}
    \label{eq:CONV}
\end{equation}
where \textit{$\textbf{W}$}, \textit{$\textbf{I}$}, \textit{$\textbf{O}$}, and \textit{$\textbf{B}$} denote the weights, input activations, output activations, and biases, respectively. In the CONV layers, $C$ and $M$, $E$ and $F$, $R$ and $S$, and $U$ stand for the number of input and output channels, the size of input and output feature maps, and the size of weight filters, and stride, respectively; while in the FC layers, $C$ and $M$ represent the number of input and output neurons, respectively; with $\sigma$ denoting the activation function, e.g., a $ReLU$ function ($ReLU(x)=max(x,0)$). The pooling layers reduce the dimension of feature maps via average or max pooling. The recently emerging compact DNNs (e.g., MobileNet~\cite{howard2017mobilenets} and EfficientNet~\cite{tan2019efficientnet}) introduce depth-wise CONV layers and squeeze-and-excite layers which can be expressed in the above description as well~\cite{chen2019eyeriss}.

\subsection{Demands for Model Compression}
During DNN inference, the weight parameters often dominate the memory storage and limit the energy efficiency due to their associated data movements \cite{PredictiveNet,Xu_2020}. 
In response, there are three main streams model compression techniques: pruning/sparsification, weight decomposition, and quantizaition.

\textbf{Pruning/sparsification.} Pruning, or weight sparsification, increases the sparsity in the weights of DNNs by zeroing out non-significant ones, which is usually interleaved with fine-tuning phases to recover the performance in practice. An intuitive method is to elementwisely zero out weights with near-zero magnitudes \cite{han2015learning}. Recent works establish more advanced pruning methods to enforce structured sparsity for convolutional layers \cite{network_slimming,li2016pruning,luo2017thinet,he2017channel}.
The
work in \cite{mao2017exploring} exhibits that due to the encoding index overhead, vector-wise sparsity is able to obtain similar compression rates at the same accuracy as element-wise/unstructured sparsity. 

\textbf{Weight decomposition. } Another type of approaches to compress DNNs is weight decompositions, e.g., low-rank decomposition. This type of compression models the redundancy in DNNs as correlations between the highly structured filters/columns in convolutional or fully connected layers \cite{RDSEC,denton2014exploiting,jin2014flattened,nakkiran2015compressing}. Low-rank decomposition expresses the highly structured filters/weights using products of two small matrices. 

\textbf{Quantization. } Quantization attempts to reduce the bit width of the data flowing through DNNs \cite{wu2016quantized,han2015deep,wang2019e2}, thus is able to shrink the model size for memory savings and simplify the operations for more efficient acceleration. In addition, it has been shown that combinations of low-rank decomposition and pruning  can lead to a higher compression ratio while preserving the accuracy \cite{Yu_2017_CVPR,gui2019adversarially}. 

\subsection{Motivation for \textit{SmartExchange}}


\begin{scriptsize}
\begin{table}[!t]
  \centering
  \caption{Unit energy cost per 8-bit extracted from a commercial 28nm technology.}
    \vspace{-0.5em}
  \label{table:formatting}
  \begin{tabular}{|c|c|c|c|c|c|}
    \hline
     & DRAM & SRAM & MAC & multiplier & adder \\
    \hline
    Energy  & \multirow{2}{*}{100} & \multirow{2}{*}{1.36$-$2.45} &  \multirow{2}{*}{0.143} &  \multirow{2}{*}{0.124} & \multirow{2}{*}{0.019} \\
    (pJ/8bit) & & & & & \\
    \hline
  \end{tabular}
    \vspace{-1.2em}
  \label{table:unit_energy}
\end{table}
\end{scriptsize}

Table~\ref{table:unit_energy} shows the unit energy cost of accessing different-level memories with different storage capacities and computing an MAC/multiplication/addition (the main computation operation in DNNs) designed in a commercial 28nm CMOS technology. We can see that the unit energy cost of memory accesses is much higher ($\geq 9.5\times$) than that of the corresponding MAC computation. Therefore, it is promising in terms of more efficient acceleration if we can potentially enforce higher-order of weight structures to more aggressively trade higher-cost memory accesses for lower-cost computations, motivating our \textit{SmartExchange} decomposition idea. That is, the resulting higher structures in DNN weights' decomposed matrices, e.g., $C_e$ in Figure~\ref{fig:SE_concept}, will enable much reduced memory accesses at a cost of more computation operations (i.e., shift-and-add operations in our design), as compared to the vanilla networks. 

In addition, the integration of decomposition, pruning, and quantization, i.e., our \textit{SmartExchange}, is motivated by the hypothesis of potentially \textit{higher-order sparse structures} as recently observed in~\cite{yu2017compressing, gui2019model} from an algorithm perspective. That is, rather than enforcing element-wise sparsity on the original weight matrix directly, it is often more effective to do so on corresponding decomposed matrix factors (either additive or multiplicative). 
Note that \textit{SmartExchange} on the algorithm level targets a more hardware favorable weight representation, and thus can be combined with other activation representations (e.g., sparse activations)~\cite{parashar2017scnn, zhou2018cambricon, judd2016stripes, albericio2017bit, delmas2019bit} for maximizing the efficiency gains.

To summarize, the overall goal of \textit{SmartExchange} is to trade higher-cost data movement/access for lower-cost weight reconstruction (MACs or shift-and-add operations). To achieve this goal, the concrete design of \textit{SmartExchange} is motivated from the following two folds: 1) seeking more compactness in the weight representation (contributed mainly by \textbf{sparsity} and also by the \textbf{decomposition} which might help discover higher-order sparse structures); and 2) reducing the multiplication workload in the weight reconstruction (contributed mainly by the special power-of-two \textbf{quantization} of nonzero elements).

\section{The Proposed SmartExchange Algorithm}\label{sec:algorithm}

In this section, we first formulate the \textit{SmartExchange} decomposition problem. To our best knowledge, \textit{SmartExchange} algorithm is the first unified formulation that conceptually combines three common methodologies for compressing and speeding up DNNs: weight \textit{sparsification} or pruning, weight matrix \textit{decomposition}, and weight \textit{quantization}. We then develop an efficient algorithm to solve the problem, and show that the \textit{SmartExchange} algorithm can be appended to post-processing a trained DNN for compression/acceleration without compromising the accuracy loss. We then demonstrate that \textit{SmartExchange} algorithm can be incorporated into DNN re-training to achieve more promising trade-offs between the inference accuracy and resource usage (e.g., the model size, memory access energy, and computational cost). 

\subsection{Problem Formulation}

Previous works have tried to compress DNNs by reducing the correlation between filters (in CONV layers) or columns (in FC layers) via decomposing weights \cite{denton2014exploiting,jin2014flattened,nakkiran2015compressing}. Here, given a weight matrix $W\in\mathbb{R}^{m\times n}$,
we seek to decompose it as the product of a coefficient matrix $C_e\in\mathbb{R}^{m\times r}$ and a basis matrix $B\in\mathbb{R}^{r\times n}$ where $r \le \min\{m,n\}$, such that
\begin{align}
    W \approx C_e B 
    \label{eq:rank}
\end{align}

In addition to suppressing the reconstruction error (often defined as $||W-C_e B||_F^2$), we expect the decomposed matrix factors to display more favorable structures for compression/acceleration. In the decomposition practice, $B$ is usually constructed to be a very small matrix (e.g., $B$ takes the values of 3, 5, 7, whereas $C_e$ has $m$ rows with $m$ being the number of weight vectors in a layer). For the much larger $C_e$, we \textbf{enforce the following two structures simultaneously} to aggressively boost the energy efficiency: 1) $C_e$ needs to be highly sparse (a typical goal of pruning); and 2) the values of the non-zero elements in $C_e$ are exactly the powers of $2$, so that their bit representations can be very compact and their involved multiplications to rebuild the original weights from $B$ and $C_e$ are simplified into much lower-cost shift-and-add operations. As a result, instead of storing the whole weight matrix, the proposed \textit{SmartExchange} algorithm requires storing only a very small $B$ and a large, yet highly sparse and readily quantized $C_e$. Therefore, the proposed algorithm greatly reduces the overall memory storage, and makes it possible to hold $C_e$ in a much smaller memory of a lower-level memory hierarchy to minimize data movement costs. We call such \{$C_e$, $B$\} pair the \textit{SmartExchange form} of $W$.

The rationale of the above setting arises from previous observations of composing pruning, decomposition, and quantization. For example, combining matrix decomposition and pruning has been found to effectively compress the model without notable performance loss \cite{jin2014flattened,nakkiran2015compressing,gui2019model}. One of our innovative assumptions is to require non-zero elements to take one of a few pre-defined discrete values, that are specifically picked for not only compact representations but also lower-cost multiplications. Note that it is different from previous DNN compression using weight clustering, whose quantized values are learned from data \cite{gong2014compressing,wu2018deep}. 

\textit{SmartExchange} decomposition problem can hence be written as a constrained optimization:
\begin{align}
    \underset{C_e,B}{\mathrm{arg}\:}{ \mathrm{min}} & \quad \|W - C_e B\|_F^2    
        \label{eq:sed}
    \\
    \mathrm{s.t.} &\quad \sum_j\ \|C_e[:,j]\|_0 \le S_c, \ \ {C_e}[i,j]\in\Omega_P,\ \ \forall i,j \ \ |P| \leq N_p, 
    \nonumber
\end{align}
where $\Omega_P\coloneqq\{0, \pm 2^p | p\in P\}$ with $P$ being an integer set whose cardinality $|P|$ is no more than $N_p$, $S_c$ controls the total number of non-zero elements in $C_e$, while $N_p$ controls the bit-width required to represent a nonzero element in $C_e$.

\subsection{The SmartExchange Algorithm}

Solving Eq. (\ref{eq:sed}) is in general intractable due to both the nonconvex $\ell_0$ constraint, and the integer set $\Omega_P$ constraint. We introduce an efficient heuristic algorithm that iterates between objective fitting and feasible set projection. The general outline of the \textit{SmartExchange} algorithm is described in Algorithm 1, and the three key steps to be iterated are discussed below:


\begin{figure}
\vspace{-1em}
    \begin{minipage}{0.47\textwidth}
        \begin{algorithm}[H]
            \caption{\textit{SmartExchange} Algorithm.}
            \label{alg:general}
            \begin{algorithmic}[1]
                \STATE{Sparsify $C_e$ in a channel-wise manner;}
                    \STATE{Initialize $C_e$ and $B$; Iteration = 0;}
                    \STATE{while 
                    \textit{$\|\delta(C_e)\| \geq tol$} or 
                    \textit{ iteration $<$ tol\_maximum}:}
                    \STATE{\hspace{10pt} \textbf{Step 1:} Quantize $C_e$ to powers of $2$;}
                    \STATE{\hspace{10pt} \textbf{Step 2:} Fitting $B$ and $C_e$;}
                    \STATE{\hspace{10pt} iteration = iteration + 1;}
                    \STATE{\hspace{10pt} \textbf{Step 3:} Sparsity $C_e$ in a vector-wise manner;}
                    \STATE{Re-quantize $C_e$ and re-fit $B$.}
            \end{algorithmic}
        \end{algorithm}
    \end{minipage}
\vspace{-1.5em}
\end{figure}


\textbf{Step 1: Quantizing $C_e$.} The quantization step projects the nonzero elements in $C_e$ to $\Omega_P$. Specifically, we will first normalize each column in $C_e$ to have a unit norm in order to avoid scale ambiguity. We will then round each non-zero element to its nearest power-of-two value. We define $\delta(C_e)$ to be the quantization difference of $C_e$. 

\textbf{Step 2: Fitting $B$ and $C_e$.} We will first fit $B$ by solving
$\mathrm{arg}\:\mathrm{min}_B  \|W - C_e B\|_F^2$,
and then fit $C_e$  by solving
$\mathrm{arg}\: \mathrm{min}_{\,C_e}  \|W - C_e B\|_F^2$.
When fitting either one, the other is fixed to be its current updated value. The step simply deals with two unconstrained least squares.

\textbf{Step 3: Sparsifying $C_e$.} To pursue better compression/acceleration, we simultaneously introduce both \textit{channel-wise} and \textit{vector-wise} sparsity to $C_e$:
\begin{itemize}
    \item We first prune channels whose corresponding scaling factor in batch normalization layers is lower than a threshold which is manually controlled for each layer. In practice, we only apply channel-wise sparsifying at the first training epoch once, given the observation that the pruned channel structure will not change much. 
    \item We then zero out elements in $C_e$ based on the magnitudes to meet the vector-wise sparsity constraint: $\sum_j\ \|C_e[:,j]\|_0 \le S_c$, where $S_c$ is manually controlled per layer. 
\end{itemize}

In practice, we use hard thresholds for channel and vector-wise sparsity to zero out small magnitudes in $C_e$ for implementation convenience.
With the combined channel and vector-wise sparsity, we can bypass reading the regions of the input feature map that correspond to the pruned parameters, saving both storage-access and computation costs in \textit{convolution operations}. Meanwhile, the sparsity patterns in $C_e$ also reduce the encoding overheads, as well as the storage-access and computation costs during the \textit{weight reconstruction} $W = C_e B$.

After iterating between the above three steps (quantization, fitting and sparsification) for sufficient iterations, we conclude the iterations by re-quantizing the nonzero elements in $C_e$ to ensure ${C_e}[i,j]\in\Omega_P$ and then re-fitting $B$ with the updated $C_e$. 

\subsection{Applying the SmartExchange Algorithm to DNNs}
\label{subsec:sed2dnn}

\textbf{\textit{SmartExchange} algorithm as post-processing.} 
The value of $r$ (see Eq. (\ref{eq:rank})) is a design knob of \textit{SmartExchange} for trading-off the achieved \textit{compression rate} and model accuracy, i.e., a smaller $r$ favors a higher compression rate yet might cause a higher accuracy loss. Note that
$r$ is equal to the rank of the basis matrix $B$, i.e., $r=n$ when $B$ is a full matrix, otherwise $r\leq n$.
To minimize the memory storage, we set the basis matrix $B\in\mathbb{R}^{r\times n}$ to be small. 
In practice, we choose $n=R=S$ with $R\times S$ being the CONV kernel size. Since $n$ is small, we choose $r = n = S$ too. 
We next discuss applying the proposed algorithm to the FC and CONV layers. 
In all experiments, we initialize $C_e=W$ and $B=I$ for simplicity.
\begin{itemize}
    \item \underline{\textit{\textit{SmartExchange} algorithm on FC layers.}} Consider a fully-connected layer $W \in R^{M \times C}$. We reshape each row of $W$ into a new matrix $\tilde{W_i}\in R^{C/S \times S}$, and then apply \textit{SmartExchange} algorithm. Specifically, zeros are padded if $C$ is not divisible by $S$, and 
    \textit{SmartExchange} algorithm is applied to $\tilde{W_i}$, where $i=1,\dots,M$. When $C\gg S$, the reconstruction error might tend to be large due to the imbalanced dimensions. We alleviate it by slicing $\tilde{W_i}$ into smaller matrices along the first dimension.
    \item \underline{\textit{\textit{SmartExchange} algorithm on CONV layers.}} Consider a convolutional layer $W$ in the shape $(M, C, R, S)$: \underline{Case 1}: $R=S>1$. We reshape the $M$ filters in $W$ into matrices of shape $(S \times C, S)$, on which \textit{SmartExchange} algorithm is applied. The matrices can be sliced into smaller matrices along the first dimension if $S \times C\gg S$. \underline{Case 2}: $R=S=1$. The weight is reshaped into a shape of $(M,C)$ and then is treated the same as an FC layer.
\end{itemize} 
The above procedures are easily parallelized along the axis of the output channels for acceleration.

We apply the \textit{SmartExchange} algorithm on a VGG19 network\footnote{https://github.com/chengyangfu/pytorch-vgg-cifar10} pre-trained on the CIFAR-10 \cite{cifar10}, with $\theta=4\PLH10^{-3}$, $tol=10^{-10}$, and a maximum iteration of 30. Weights in it are decomposed by \textit{SmartExchange} algorithm into the coefficient matrices and basis matrices. It only takes about 30 seconds to perform the algorithm on the network. Without re-training, the accuracy drop in the validation set is as small as $3.21\%$ with an overall compression rate of over 10$\times$. The \textit{overall compression rate} of a network is defined as the ratio between the total number of bits to store the weights (including the coefficient matrix $C_e$, basis matrix $B$, and encoding overhead) and the number of bits to store the original FP32 weights.

\textbf{\textit{SmartExchange} algorithm with re-training.} After a DNN has been post-processed by \textit{SmartExchange} algorithm, a re-training step can be used to remedy the accuracy drop. As the un-regularized re-training will break the desired property of coefficient matrix $C_e$, we take an empirical approach to alternate between 1) re-training the DNN for one epoch; and 2) applying the \textit{SmartExchange} algorithm to ensure the $C_e$ structure. The default iteration number is 50 for CIFAR-10 \cite{cifar10} and 25 for ImageNet \cite{Deng09imagenet}. As shown in experiments in Section \ref{subsec:compression_results}, the alternating re-training process further improves the accuracy while maintaining the favorable weight structure. More analytic solutions will be explored in future work, e.g., incorporating \textit{SmartExchange} algorithm as a regularization term \cite{wu2018deep}.

\section{The Proposed SmartExchange Accelerator}\label{sec:Dedicated_accelerator}
    \vspace{-0.3em}
In this section, we present our proposed \textit{SmartExchange} accelerator. We first introduce the design principles and considerations (Section \ref{sec:Design_challenges_principles}) for fully making use of the proposed \textit{SmartExchange} algorithm's properties to maximize energy efficiency and minimize latency, and then describe the proposed accelerator (Section \ref{subsec:accelerator}) in details.

    \vspace{-0.8em}
\subsection{Design Principles and Considerations}\label{sec:Design_challenges_principles}
    \vspace{-0.3em}
The proposed \textit{SmartExchange} algorithm exhibits a great potential in reducing the memory storage and accesses for on-device DNN inference. However, this potential cannot be fully exploited by existing accelerators~\cite{chen2014diannao, parashar2017scnn, zhang2016cambricon, albericio2017bit} due to 1) the required rebuilding operations of the \textit{SmartExchange} algorithm to restore weights and 2) the unique opportunity to explore coefficient matrices' vector-wise structured sparsity. In this subsection, we analyze the opportunities brought by the \textit{SmartExchange} algorithm to abstract design principles and considerations for developing and optimizing the dedicated \textit{SmartExchange} accelerator. 


\textbf{Minimizing overhead of rebuilding weights. }
Thanks to the sparse and readily quantized coefficient matrices resulting from the \textit{SmartExchange} algorithm, the memory storage and data movements associated with these matrices can be greatly reduced (see Table \ref{table:sed_results}; e.g., up to $80\times$). Meanwhile, 
to fully utilize the advantages of the \textit{SmartExchange} algorithm, the overhead of rebuilding weights should be minimized. To do so, it critical to ensure that the location and time of the rebuilding units and process are properly designed. Specifically, a \textit{SmartExchange} accelerator should try to 1) store the basis matrix close to the rebuild engine (RE) that restores weights using both the basis matrix and corresponding weighted coefficients; 2) place the RE to be close to or within the processing elements (PEs); and 3) use a weight-stationary dataflow for the basis matrix. Next, we elaborate these principles in the context of one 3D filter operation (see Figure \ref{fig:two_order} (a)):

\begin{figure}[!t]
    \centerline{\includegraphics[width=80mm]{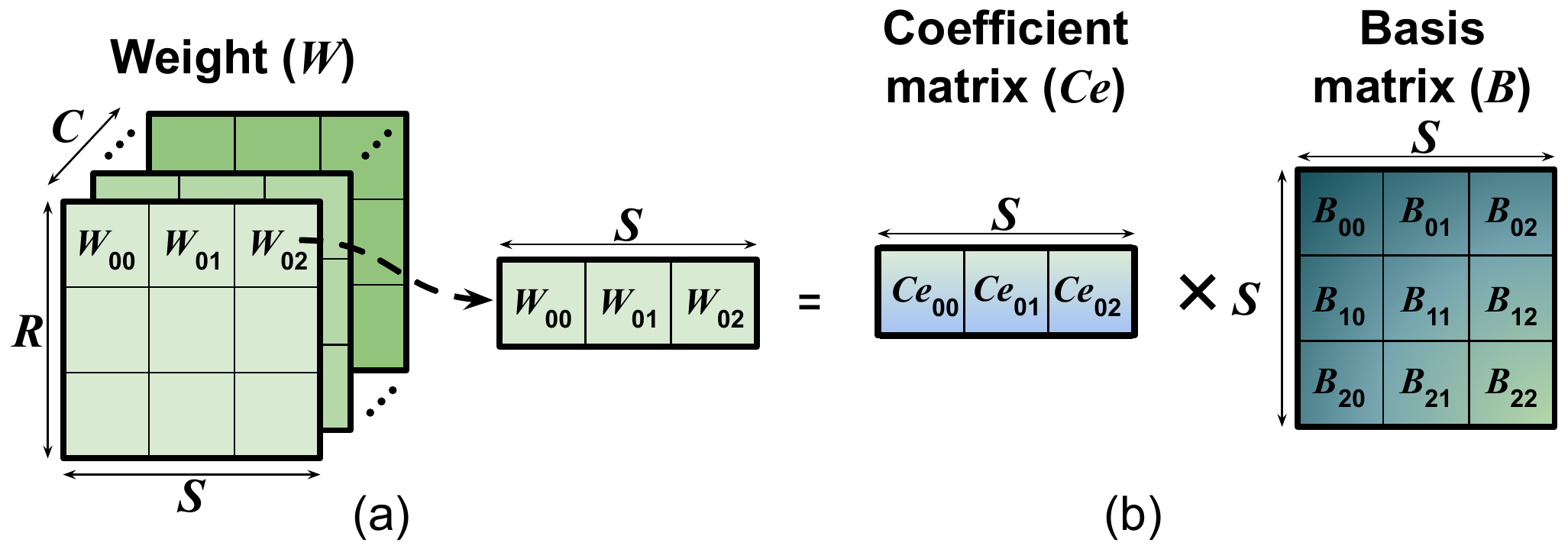}}
    \vspace{-0.5em}
    \caption{An illustration of (a) a 3D weight filter and its parameters' notations, and (b) rebuilding one row of weights using the corresponding basis and coefficient matrix.
    \label{fig:two_order}}
    \vspace{-1.8em}
\end{figure}

\ul{First}, the \textit{SmartExchange} algorithm decomposes the weight matrix (($C\times R)\times S$) corresponding to one 3D filter into a coefficient matrix of size ($C\times R)\times S$ and a basis matrix of size $S\times S$. According to Eq. (\ref{eq:rank}), each element in the basis matrix is reused $C\times R$ times in order to rebuild the weights, while the number of reuses of each element in the coefficient matrix is only $S$. This often means two orders of magnitude more reuse opportunities for the basis matrices than that of the coefficient matrices, considering most state-of-the-art DNN models. Therefore, the basis matrices should be placed close to both the PEs and REs, and stored in the local memories within REs for minimizing the associated data movement costs.

\ul{Second}, the REs should be located close to the PEs for minimizing the data movement costs of the rebuilt weights. This is because once the weights are rebuilt, the cost of their data movements are the same as the original weights. 

\ul{Third}, as the basis matrices are reused most frequently, the dataflow for these matrices should be weight stationary, i.e., once being fetched from the memories, they should stay in the PEs until all the corresponding weights are rebuilt. 

\begin{figure}[!t]
    \centerline{\includegraphics[width=83mm]{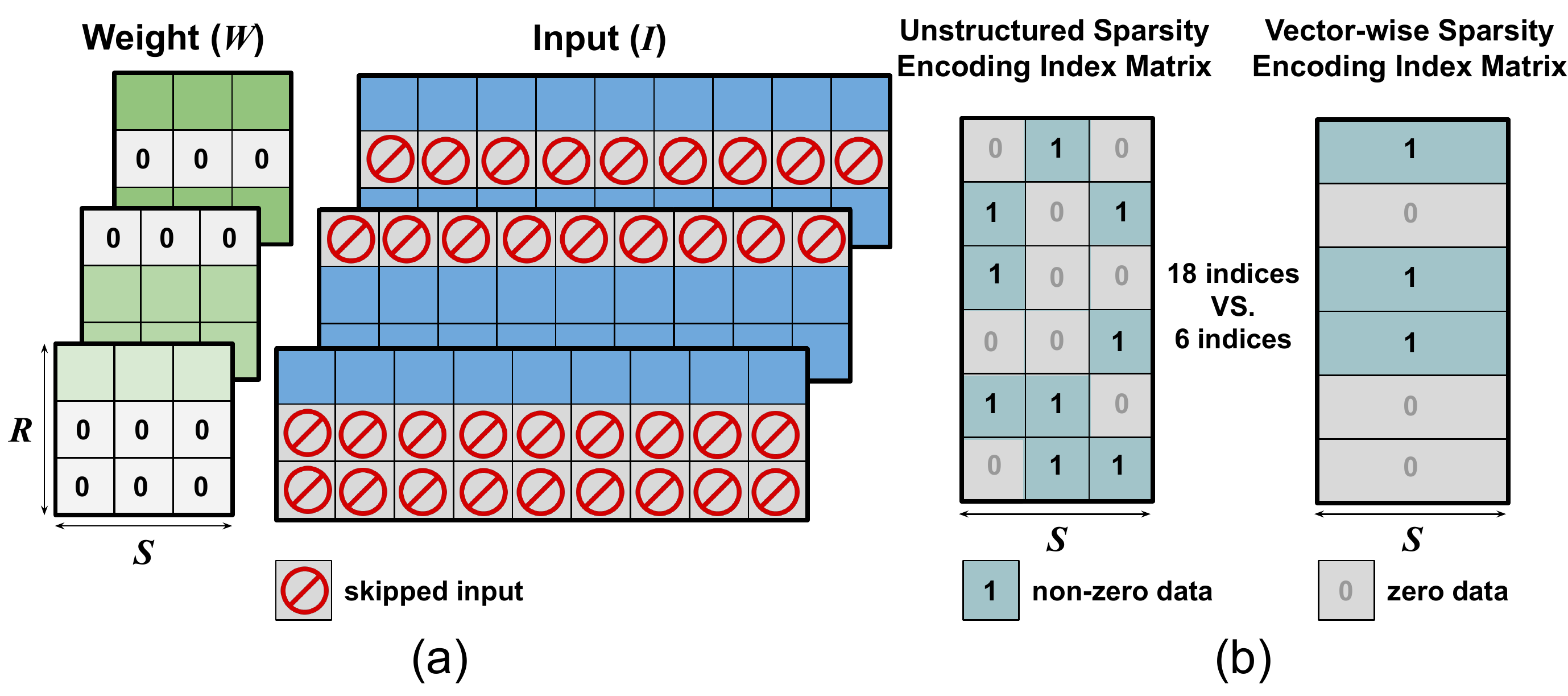}}
    \vspace{-0.8em}
    \caption{An illustration of (a) vector-wise skipping the corresponding activations, and (b) the reduced indexing overhead, thanks to the enforced vector-wise weight sparsity of the \textit{SmartExchange} algorithm.
    \label{fig:coefficient_sparsity}}
    \vspace{-1.7em}
\end{figure}

\textbf{Taking advantage of the (structured) sparsity. } The enforced vector-wise sparsity in the \textit{SmartExchange} algorithm's coefficient matrices offers benefits of 1) vector-wise skipping both the memory accesses and computations of the corresponding activations (see Figure \ref{fig:coefficient_sparsity} (a)) and 2) reduced coefficient matrix encoding overhead (see Figure \ref{fig:coefficient_sparsity} (b)). Meanwhile, there is an opportunity to make use of the vector-wise/bit-level sparsity of activations for improving efficiency.

\ul{First}, one promising benefit of the \textit{SmartExchange} algorithm's enforced vector-wise sparsity in the coefficient matrices is the possibility to vector-wise skip both the memory accesses and computations of the corresponding activations (see Figure \ref{fig:coefficient_sparsity} (a)). This is because those vector-wise sparse coefficient matrices' corresponding weight vectors naturally carry their vector-wise sparsity pattern/location, offering the opportunity to directly use the sparse coefficient matrices' encoding index to identify the weight sparsity and skip the corresponding activations' memory accesses and computations. Such a skipping can lead to large energy and latency savings because weight vectors are shared by all activations of the same fracture maps in CONV operations, see Figure \ref{fig:coefficient_sparsity} (b).

\begin{figure}[!b]
\vspace{-2em}
    \centerline{\includegraphics[width=88mm]{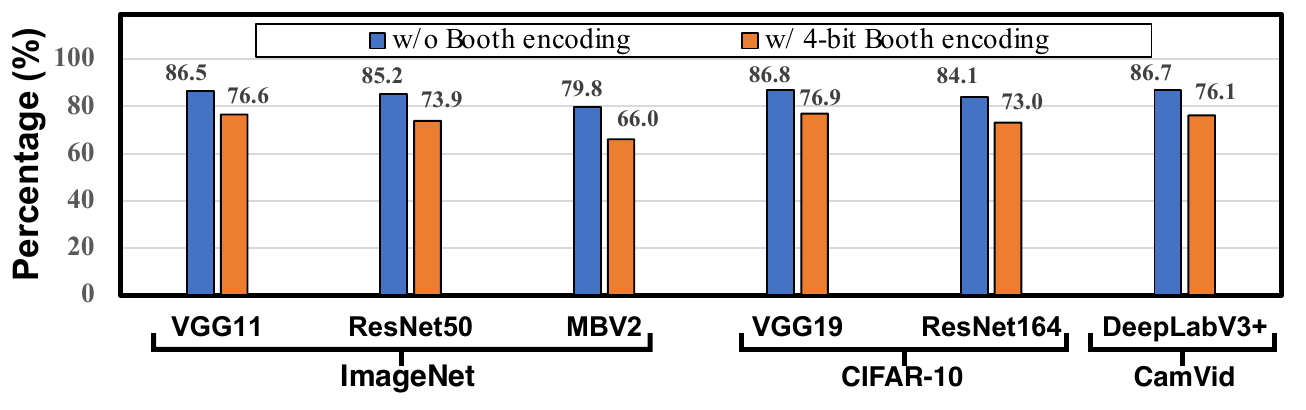}}
    \vspace{-0.8em}
    \caption{The bit-level sparsity in activations for six models on three datasets.} 
    \label{fig:activatioin_sparsity}
\end{figure}

\begin{figure*}[!t]
    \centerline{\includegraphics[width=168mm]{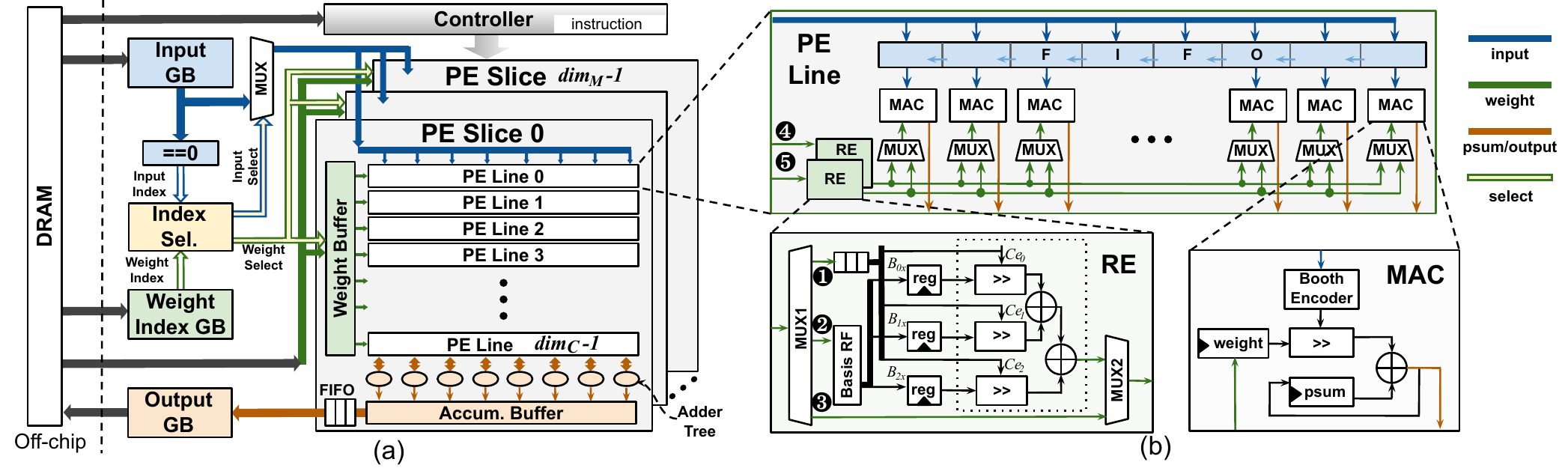}}
    \vspace{-0.8em}
    \caption{An illustration of the proposed \textit{SmartExchange} accelerator: (a) architecture, and (b) the block diagram of the processing element (PE) line, each of which includes two rebuilding engines (REs) and eight multiply-and-accumulate (MAC) units.}
    \vspace{-1.5em}
    \label{fig:overview_PE}
\end{figure*}


\ul{Second}, commonly used methods for encoding weight sparsity, such as run-length coding (RLC)~\cite{zhang2016cambricon, eyeriss}, the 1-bit direct weight indexing ~\cite{zhou2018cambricon}, and Compressed Row Storage (CRS)~\cite{han2016eie}, store both the values and sparsity encoding indexes
of weights. Our \textit{SmartExchange} algorithm's vector-wise weight sparsity reduce both the sparsity encoding overhead (see Figure \ref{fig:coefficient_sparsity} (b)) and skipping control overhead. The resulting energy and latency benefits depend on the sparsity ratio and pattern, and hardware constraints (e.g., memory bandwidths).




\ul{Third}, the accelerator can further make use of bit-level and vector-wise sparsity of activations to  improve energy efficiency and reduce latency, where the bit-/vector-wise sparsity means the percentage of the zero activation bits/rows over the total activation bits/rows. 
Figure \ref{fig:activatioin_sparsity} shows the bit-level sparsity of activations w/ and w/o 4-bit Booth encoding~\cite{delmas2019bit} in popular DNNs, including VGG11, ResNet50, and MobileNetV2 on ImageNet, VGG19 and ResNet164 on CIFAR-10, and DeepLabV3+ on CamVid. We can see that the bit-level sparsity is 79.8\% under an 8-bit precision and 66.0\% using the corresponding 4-bit Booth encoding even for a compact model like MobileNetV2; for vector-wise sparsity, it can be widely observed among the CONV layers with $3 \times 3$ kernel size, e.g., up to 27.1\% in the last several CONV layers of MobileNetV2 and up to 32.4\% in ResNet164. If the memory accesses and computations of zero activation bits can be skipped, the resulting performance improvement will be proportional to the bit-level activation sparsity, as elaborated in \cite{delmas2019bit} which shows that combining with zero weights, higher efficiency can be achieved when targeting zero activation bits (instead of merely considering zero activations). As for the vector-wise sparsity of activations, only when activations at one row are all zeros, we could skip fetching the corresponding weight vectors due to the window sliding processing of CONV layers.

\textbf{Support for compact models. } The recently emerged compact models, such as MobileNet \cite{howard2017mobilenets} and EfficentNet \cite{tan2019efficientnet}, often adopt depth-wise CONV and squeeze-and-excite layers other than the traditional 2D CONV layers to restrict the model size, which reduces the data resuse opportunities. Take a depth-wise CONV layer as an example, it has an ``extreme" small number of CONV channels (i.e., 1), reducing the input reuse over the standard CONV layers; similar to that of FC layers, there are no weight reuse opportunities in squeeze-and-excite layers. On-device efficient accelerators should consider these features of compact models for their wide adoption and leveraging compact models for more efficient processing. 



\subsection{Architecture of the SmartExchange Accelerator}\label{subsec:accelerator}
   \vspace{-0.5em}
\textbf{Architecture overview.} Figure~\ref{fig:overview_PE} (a) shows the architecture of the proposed \textit{SmartExchange} accelerator which consists of a 3D PE array with a total of $dim_M$ PE slices, input/index/output global buffers (see the blocks named Input GB, Weight Index GB, and Output GB, where GB denotes global buffer) associated with an index selector for sparsity (see the blocks named Index sel.), and an controller. The accelerator communicates with an off-chip DRAM through DMA (direct memory access) \cite{zhang2016cambricon}. Following the aforementioned design principles and considerations (see Section \ref{sec:Design_challenges_principles}), the proposed accelerator features the following properties: \ul{1) \textit{an RE design}} which is inserted within PE lines to reduce the rebuilding overhead (see the top part of Figure~\ref{fig:overview_PE} (b)); 
\ul{2) \textit{a hybrid dataflow}}: an 1D row stationary dataflow is adopted within each PE line for maximizing weight and input reuses, while each PE slice uses an output stationary dataflow for maximizing output partial sum reuses; 
\ul{3) \textit{an index selector}} (named Index Sel. in Figure~\ref{fig:overview_PE} (a)) to select the none-zero coefficient and activation vector pairs as inspired by~\cite{zhou2018cambricon}. 
This is to skip not only computations but also data movements associated with the sparse rows of the coefficients and activations. The index selector design in \textit{SmartExchange} is the same as that of \cite{zhou2018cambricon} except that here index values of 0/1 stand for vector (instead of scalar) sparsity; 
\ul{4) \textit{a data-type driven memory partition}} in order to use matched bandwidths (e.g., a bigger bandwidth for the weights/inputs and a smaller bandwidth for the outputs) for different types of data to reduce the unit energy cost of accessing the SRAMs which is used to implement the GB blocks~\cite{du2015shidiannao}. We adopt separated centralized GBs to store the inputs, outputs, weights and indexes, respectively, and distributed SRAMs (see the Weight Buffer unit in Figure~\ref{fig:overview_PE} (a)) among PE slices to store weights (including the coefficients and basis matrices); 
and \ul{5) \textit{a bit-serial multiplier based MAC array}} in each PE line to make use of the activations' bit-level sparsity together with a Booth Encoder as inspired by~\cite{delmas2019bit}.

\begin{figure}[!t]
    \centerline{\includegraphics[width=60mm]{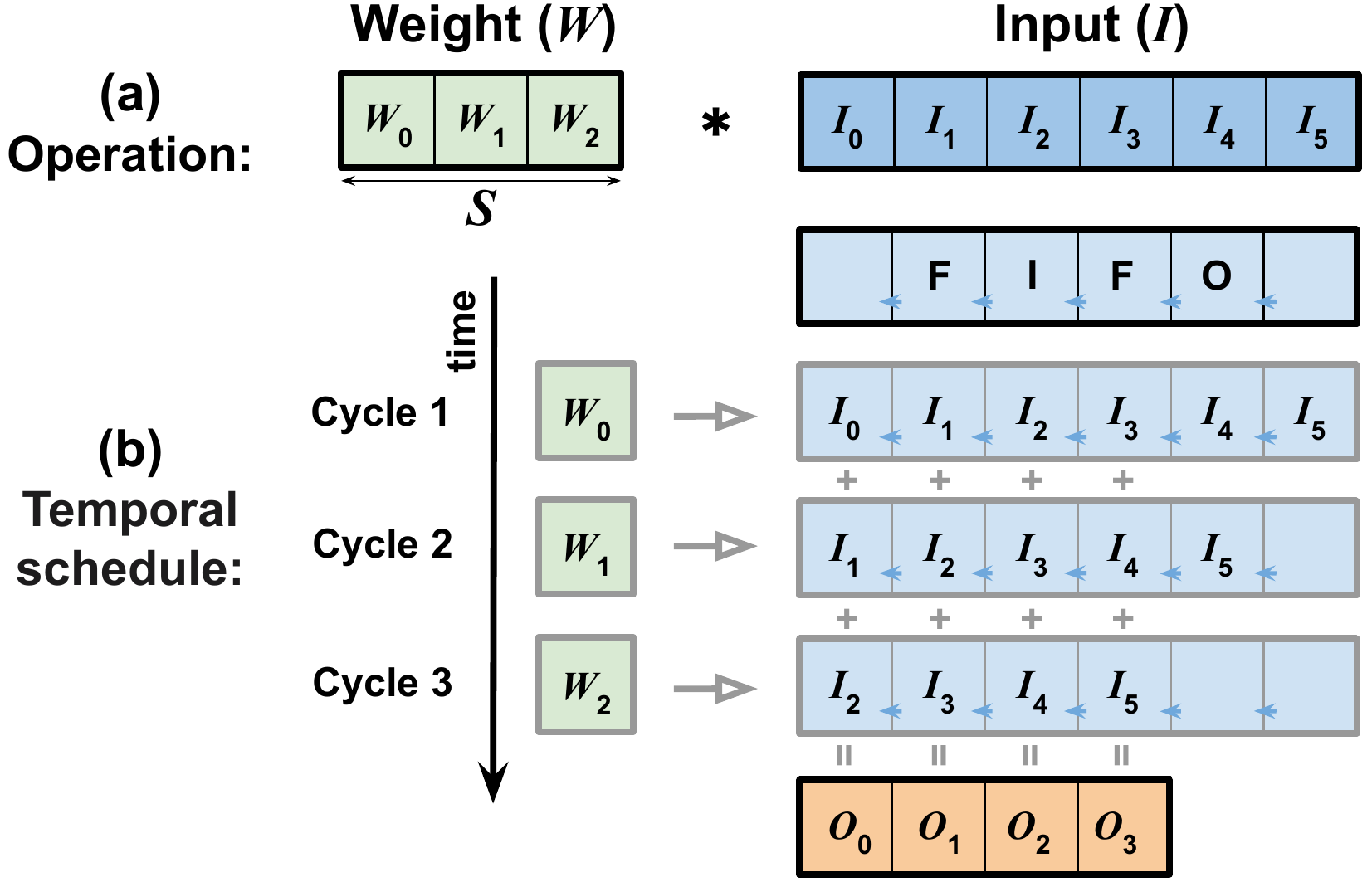}}
    \vspace{-0.5em}
    \caption{An illustration of the proposed 1D row stationary along each PE slice (in this particular example, FIFO size is 5, and in general it should be $dim_F+S-1$): (a) 1D CONV and (b) processing flow of 1D row stationary. }
    \label{fig:1D_conv}
     \vspace{-2em}
\end{figure}

\textbf{PE slices and dataflow.} We here describe the design of the PE Slice unit in the 3D PE slice array of Figure~\ref{fig:overview_PE} (a): 

\ul{First, the 3D PE Slide array:} our \textit{SmartExchange} accelerator enables paralleled processing of computations associated with the same weight filter using the PE slice array of size $dim_M$ (with each PE slice having $dim_C$ PE lines) and $dim_C$ number of input channels, where the resulting partial sums are accumulated using the adder trees at the bottom of the PE lines (see the bottom right side of Figure~\ref{fig:overview_PE} (a)). In this way, a total of $dim_M$ consecutive output channels (i.e., $dim_M$ weight filters) are processed in parallel to maximize the reuse of input activations. Note that this dataflow is employed to match the way we reshape the weights as described in Section~\ref{sec:algorithm}.C.

\ul{Second, the PE line design:} each PE line in Figure~\ref{fig:overview_PE} includes an array of $dim_F$ MACs, one FIFO (using double buffers), and two RE units, where the REs at the left restore the original weights in a row-wise manner. 
During operations, each PE line processes one or multiple 1D CONV operations, similar to the 1D row stationary in \cite{eyeriss} except that we stream each rebuild weight of one row temporally along the MACs for processing one row of input activations. 
In particular, the 1D CONV operation is performed by shifting the input activations along the array of MACs within the PE line (see Figure~\ref{fig:1D_conv}) via an FIFO; this 1D CONV computation is repeated for the remaining 1D CONV operations to complete one 2D CONV computation in $\leq (S\times R$) cycles (under the assumption of w/ sparsity and w/o bit-serial multiplication) with 1) each weight element being shared among all the MACs in each cycle, and 2) the intermediate partial sums of the 2D CONV operations are accumulated locally in each MAC unit (see the bottom right part of Figure~\ref{fig:overview_PE} (b)). 

\ul{Third, the RE design:} as shown in the bottom left corner of Figure~\ref{fig:overview_PE} (b), an RE unit includes an RF (register file) of size $S\times S$ to store one basis matrix and a shift-and-add unit to rebuild weights. The time division multiplexing unit at the left, i.e., MUX1, is to fetch the {\color{black}{\ding{182}}} coefficient matrices, {\color{black}{\ding{183}}} basis matrices, or {\color{black}{\ding{184}}} original weights. This design enables the accesses of these three types of data to be performed in a time division manner in order to reduce the weight bandwidth requirement by taking advantage of the fact that it is not necessary to fetch these three types of data simultaneously. Specifically, the basis matrix is fetched first and stored stationary within the RE until the associated computations are completed; the weights are then rebuilt in an RE where each row of a coefficient matrix stays stationary until all its associated computations are finished (see  Figure~\ref{fig:two_order}). The third path of MUX1 {\color{black}{\ding{184}}} for the original weights is to handle DNNs' layers where \textit{SmartExchange} is not applied on. 

\ul{Fourth, the handling of compact models:} when handling compact models, we consider an adjusted dataflow and PE line configuration for improving the utilization of both the PE slice array and the MAC array within each PE line. Specifically, for depth-wise CONV layers, since the number of CONV channels is only 1, the $dim_C$ PE lines will no longer correspond to input channels. Instead, we map the $R$ number of 1D CONV operations along the dimension of the weight height to these PE lines. 
For squeeze-and-excite/FC layers, each PE line's MAC array of $dim_F$ MACs can be divided into multiple clusters (e.g., two clusters for illustration in the top part of Figure~\ref{fig:overview_PE} (b)) with the help of the two REs in one PE line (denoted as {\color{black}{\ding{185}}} and {\color{black}{\ding{186}}}) and multiplexing units at the bottom of the MAC array, where each cluster handles computations corresponding to a different output pixel in order to improve the MAC array's utilization and thus latency performance. 
In this way, the proposed \textit{SmartExchange} accelerator's advantage is maintained even for compact models, thanks to this adjustment together with 1) our adopted 1D row stationary dataflow within PE lines, 2) the employed bit-serial multipliers, and 3) the possibility to heavily quantized coefficients (e.g., 4-bit).

\textbf{Coefficient matrix indexing.}\label{sec:encoding}
For encoding the sparse coefficients, there are two commonly used methods: 1) a 1-bit direct indexing where the indexes are coded with 1-bit (0 or 1 for zero or non-zero coefficients, respectively) \cite{zhou2018cambricon}; and 2) an RLC indexing for the number of zero coefficient rows \cite{eyeriss}. Since \textit{SmartExchange} algorithm (see Section~\ref{sec:algorithm}) enforces channel-wise sparsity first and then vector-wise sparsity on top of channel-wise sparsity, the resulting zero coefficients are mostly clustered within some regions. As a result, a 1-bit direct indexing can be more efficient with those clustered zero coefficients removed.




\textbf{Buffer design.}\label{sec:GB} 
For making use of DNNs' (filter-/vector-wise or bit-level) sparsity for skipping corresponding computations/memory-accesses, it in general requires a larger buffer (than that of corresponding dense models) due to the unknown dynamic sparsity patterns. We here discuss how we balance between the skipping convenience and the increased buffer size. Specifically, to enable the processing with sparsity, the row pairs of non-zero input activations and coefficients are selected from the Input GB and the Index GB (using the corresponding coefficient indexes), respectively, as inspired by~\cite{zhou2018cambricon}, which are then sent to the corresponding PE lines for processing with the resulting outputs being collected to the output GB. 


\ul{First, input GB:} to ensure a high utilization of the PE array, a vanilla design requires $(dim_C \times dim_F \times bits_{input})\times$ input activation rows (than that of the dense model counterpart) to be fetched for dealing with the dynamic sparsity patterns, resulting in $(dim_C \times dim_F \times bits_{input})\times$ increased input GB bandwidth requirement. In contrast, our design leads to a $\geq 1/S$ reduction of this required input GB bandwidth, with $dim_C \times dim_F \times bits_{input}$ inputs for every ($S$ + ``Booth encoded non-zero activation bits") cycles. This is because all the FIFOs in the PE lines are implemented in a ping-pong manner using double buffers, thanks to the fact that 1) the adopted 1-D row stationary dataflow at each PE line helps to relief this bandwidth requirement, because each input activation row can be reused for $S$ cycles; and 2) the bit-serial multipliers takes $\geq1$ cycles to finish an element-wise multiplication.



\ul{Second, weight/index/output buffer:} Similar to that of the input GB, weight/index buffer bandwidth needs to be expanded for handling activation sparsity, of which the expansion is often small thanks to the common observation that the vector-wise activation sparsity ratio is often relatively low. 
Note that because basis matrices need to be fetched and stored into the RE before the fetching of coefficient matrices and the weight reconstruction computation, computation stalls occur if the next basis matrix is fetched after finishing the coefficient fetching and the computation corresponding to the current basis matrix. 
Therefore, we leverage the two REs ({\color{black}{\ding{185}}} and {\color{black}{\ding{186}}} paths) in each PE line to operate in a ``ping-pong" manner to avoid the aforementioned computation stalls. 
For handling the output data, we adopt an FIFO to buffer the outputs from each PE slice before writing them back into the GB, i.e., a cache between the PE array and the output GB. This is to reduce the required output GB bandwidth by making use of the fact that each output is calculated over several clock cycles.

\begin{figure}[!t]
    \centerline{\includegraphics[width=80mm]{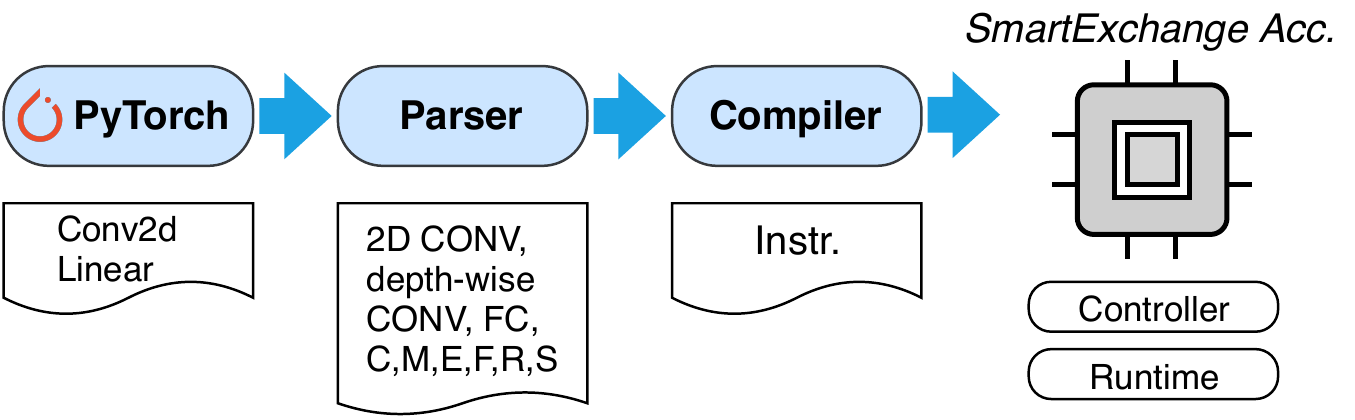}}
    \vspace{-0.5em}
    \caption{The software-hardware interface pipeline for the proposed \textit{SmartExchange} accelerator. }
    \label{fig:interface}
    \vspace{-1.5em}
\end{figure}

\textbf{Software-hardware interface.}\label{sec:interface}
Here we briefly describe how the software-hardware interface works for deploying a \textit{SmartExchange} algorithm-based DNN model from deep learning frameworks (e.g., PyTorch) into the \textit{SmartExchange} accelerator hardware. As shown in Figure \ref{fig:interface}, a pre-trained \textit{SmartExchange} algorithm-based DNN model will pass through the blocks of DNN Parser and Compiler before being loaded into the accelerator. Specifically, the DNN Parser firstly helps to extracts DNN model parameters including layer type (e.g., 2D CONV, depth-wise CONV, or FC layer) and activation and weight dimensions, which will then used by the DNN Compiler to 1) determine the dataflow and 2) generate the sparse index and instructions for configuring the PE array, memory data arrangements, and runtime scheduling. 
Finally, the resulting instructions from the Compiler are loaded into the accelerator's controller for controlling processing.

\section{Experiment Results}\label{sec:experiment}
In this section, we present a thorough evaluation of \textit{SmartExchange}, a new algorithm (see Section \ref{sec:algorithm}) and hardware (see Section \ref{sec:Dedicated_accelerator}) co-design framework.

\ul{On the algorithm level}, as \textit{SmartExchange} unifies three mainstream model compression ideas: \textit{sparsification/pruning}, \textit{decomposition}, and \textit{quantization} into one framework, we perform extensive ablation studies (benchmark over two structured pruning and four quantization, i.e., state-of-the-art compression techniques on four standard DNN models with two datasets) to validate its superiority. In addition, we evaluate \textit{SmartExchange} on two compact DNN models (MobileNetV2~\cite{sandler2018mobilenetv2} and EfficientNet-B0~\cite{tan2019efficientnet}) on the ImageNet \cite{Deng09imagenet} dataset, one segmentation model (DeepLabv3+~\cite{chen2017deeplab}) on the CamVid \cite{brostow2008segmentation} dataset, and two MLP models on MNIST.

\ul{On the hardware level}, as the goal of the proposed \textit{SmartExchange} is to boost hardware acceleration energy efficiency and speed, we evaluate \textit{SmartExchange}'s algorithm-hardware co-design results with state-of-the-art DNN accelerators in terms of energy consumption and latency when processing representative DNN models and benchmark datasets. Furthermore, to provide more insights about the proposed \textit{SmartExchange}, we perform various ablation studies to visualize and validate the effectiveness of \textit{SmartExchange}'s component techniques.


\begin{figure}[!t]
    \centerline{\includegraphics[width=90mm]{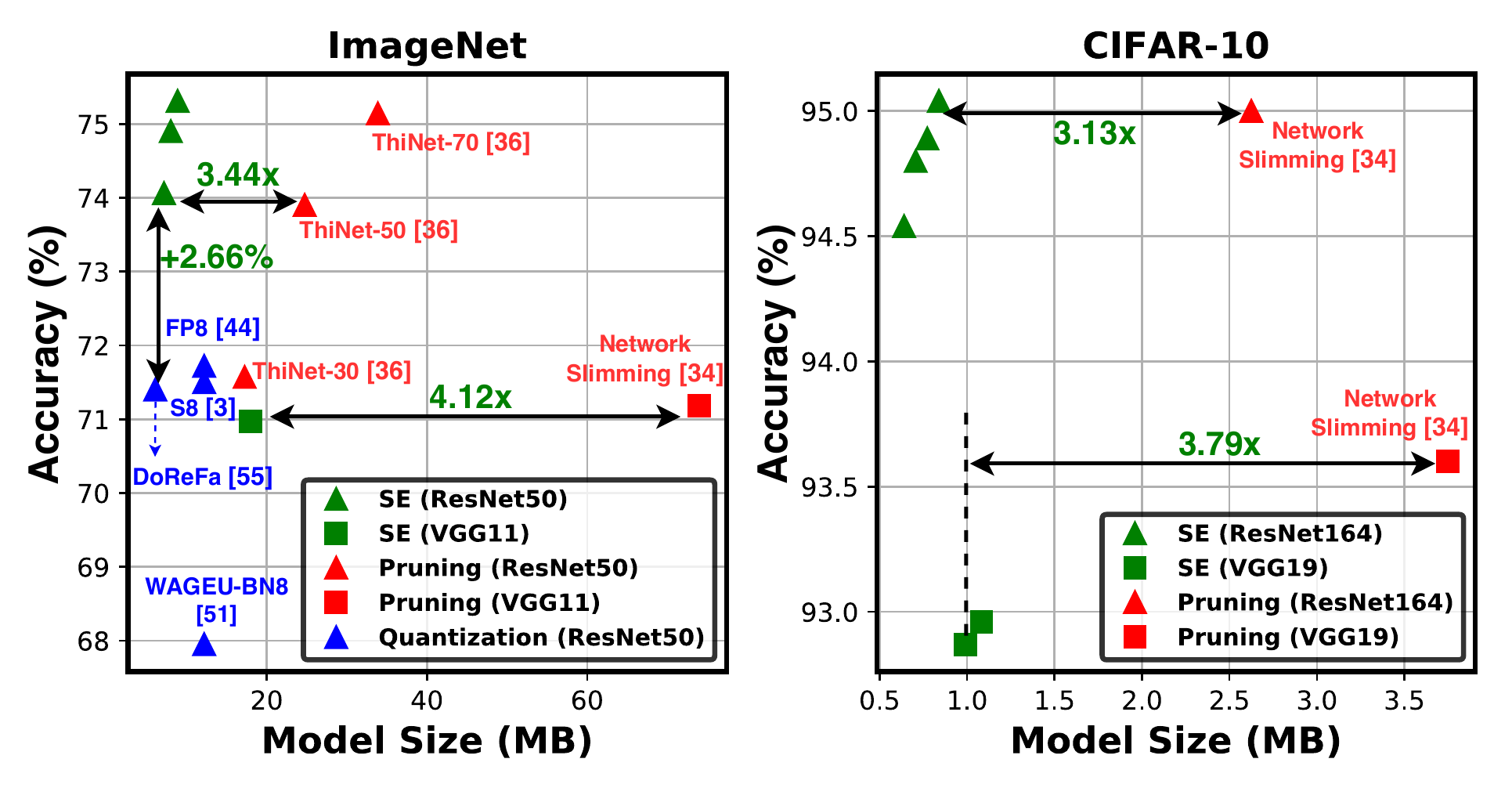}}
    \vspace{-0.8em}
    \caption{Accuracy vs. model size comparison of the \textit{SmartExchange} algorithm (SE) and state-of-the-art compression techniques on the (a) ImageNet and (b) CIFAR-10 datasets, where different colors differentiate the SE and baseline techniques.}
     \vspace{-1.8em}
    \label{fig:comparison}
\end{figure}

\vspace{-0.5em}
\subsection{Evaluation of the \textit{SmartExchange} Algorithm }
\label{subsec:compression_results}
\vspace{-0.3em}

\textbf{Experiment settings.} 
To evaluate the algorithm performance of \textit{SmartExchange}, we conduct experiments on 1) a total of \textbf{six DNN models} using both the CIFAR-10 \cite{cifar10} and ImageNet \cite{Deng09imagenet} datasets, 2) \textbf{one segmentation model} on the CamVid \cite{brostow2008segmentation} dataset, and 3) \textbf{two MLP models} on the MNIST dataset and compare the performance with state-of-the-art compression techniques in terms of accuracy and model size, including two structured pruning techniques (Network Slimming \cite{network_slimming} and ThiNet\cite{thinet}), four quantization techniques (Scalable 8-bit (S8)\cite{banner2018scalable}, FP8 \cite{fp8}, WAGEUBN \cite{WAGEUBN}, and DoReFa \cite{zhou2016dorefa}), one power-of-two quantization technique \cite{qin2019accelerating}, and one pruning and quantization technique \cite{zhou2018cambricon}.

\begin{scriptsize}
\begin{table}[!t]
    \centering
    \caption{The result summary of the proposed \textit{SmartExchange} with re-training on: 1) VGG11 and ResNet50 using the ImageNet dataset \cite{Deng09imagenet}; 2) VGG19 and ResNet164 using the CIFAR-10 dataset \cite{cifar10}; and 3) MLP-1\cite{qin2019accelerating} and MLP-2\cite{zhou2018cambricon} using the MNIST dataset.}
    \vspace{-0.8em}
    \small{
    \begin{tabular}{p{40pt} p{23pt} p{23pt} p{12pt} p{20pt} p{12pt} p{12pt} p{18pt}} \toprule
    \multirow{2}{*}{Model} & Top-1 & Top-5 & CR & Param. & $B$  & $C_e$ & Spar.  \\
                           & (\%)  & (\%)  &  ($\times$)  &  (MB)  & (MB) & (MB)  & (\%)  \\
    \midrule 

    VGG11        & 71.18\% & 90.08\% & -             & 845.75 & -    & -     & - \\ 
    $\mathrm{VGG11_{SE}}$   & 70.97\% & 89.88\% & 47.04 & 17.98  & 1.67 & 14.77 & 86.00 \\ \midrule



    ResNet50      & 76.13\%   & 92.86\%   & -           & 102.40 & -    & -    & - \\
    $\mathrm{ResNet50_{SE}}$ & 75.31\%   & 92.33\%   & 11.53  & 8.88 & 1.40 & 6.77 & 45.00 \\ 
    $\mathrm{ResNet50_{SE}}$ & 74.06\%   & 91.53\%   & 14.24  & 7.19 & 1.40 & 5.08 & 58.60 \\
    \bottomrule
    VGG19 & 93.66\%   & -  & - & 80.13 & - & - & - \\
    $\mathrm{VGG19_{SE}}$ & 92.96\%   & -   & 74.19  & 1.08 & 0.27 & 0.74 & 92.80 \\
    $\mathrm{VGG19_{SE}}$ & 92.87\%   & -   & 80.94  & 0.99 & 0.27 & 0.65 & 93.70 \\
    \midrule
    ResNet164 & 94.58\%   & -   & - & 6.75 & - & - & - \\
    $\mathrm{ResNet164_{SE}}$ & 95.04\%   & -   & 8.04  & 0.84 & 0.25 & 0.53 & 37.60 \\
    $\mathrm{ResNet164_{SE}}$ & 94.54\%   & -   & 10.55  & 0.64 & 0.25 & 0.33 & 61.00 \\
    \midrule
    MLP-1 & 98.47\%   & -   & - & 14.125 & - & - & - \\
    MLP-$\mathrm{1_{SE}}$ & 97.32\%   & -   & 130  & 0.11 & 0.01 & 0.10 & 82.34 \\
    \midrule
    MLP-2 & 98.50\%   & -   & - & 1.07 & - & - & - \\
    MLP-$\mathrm{2_{SE}}$ & 98.11\%   & -   & 45.03  & 0.024 & 0.00 & 0.024 & 93.33 \\
        \bottomrule
    \end{tabular}
    }
    \begin{tablenotes}
    \item{1. The baseline models use 32-bit floating-point representations for the weights and input/output activations, so as to benchmark with the best achievable accuracy results in the literature.}
    \item{2. The proposed \textit{SmartExchange} models use 8-bit fixed-point representations for the input/output activations; and 4-bit/8-bit representations for the coefficient/basis matrices, respectively.}
    \end{tablenotes}
     \vspace{-1.5em}
    \label{table:sed_results}
\end{table}
\end{scriptsize}

\textbf{\textit{SmartExchange} vs. existing compression techniques.} 
As \textit{SmartExchange} unifies the three mainstream ideas of pruning, decomposition and quantization, we evaluate the \textit{SmartExchange} algorithm performance by comparing it with state-of-the-art pruning-alone and quantization-alone algorithms\footnote{we did not include decomposition-alone algorithms since their results are not as competitive and also less popular.}, under four DNN models and two datasets. The experiment results are shown in Figure \ref{fig:comparison}. \textit{SmartExchange} in general outperforms all other pruning-alone or quantization-alone competitors, in terms of the achievable trade-off between the accuracy and the model size. Taking ResNet50 on ImageNet as an example, the quantization algorithm DoReFa\cite{zhou2016dorefa} seems to aggressively shrink the model size yet unfortunately cause a larger accuracy drop; while the pruning algorithm ThiNet\cite{thinet} maintains competitive accuracy at the cost of larger models. In comparison, \textit{SmartExchange} combines the best of both worlds: it obtains almost as high accuracy as the pruning-only ThiNet\cite{thinet}, which is 2.66\% higher than the quantized-only DoReFa \cite{zhou2016dorefa}; and on the other hand, it keeps the model as compact as DoReFa\cite{zhou2016dorefa}.
Apart from the aforementioned quantization works, we also evaluate the \textit{SmartExchange} algorithm with a state-of-the-art power-of-two quantization algorithm \cite{qin2019accelerating} based on the same MLP model with a precision of 8 bits: when having a higher compression rate of 130$\times$ (vs. 128$\times$ in \cite{qin2019accelerating}), \textit{SmartExchange} achieves a comparable accuracy (97.32\% vs. 97.35\%), even if \textit{SmartExchange} is not specifically dedicated for FC layers while the power-of-two quantization \cite{qin2019accelerating} does. In addition, 
compared with the pruned and quantized MLP model in \cite{zhou2018cambricon}, \textit{SmartExchange} achieves a higher compression rate of 45.03$\times$ (vs. 40$\times$ in \cite{zhou2018cambricon}) with a comparable accuracy (98.11\% vs. 98.42\%).

\begin{scriptsize}
\begin{table}[!t]
    \centering
    \caption{Evaluation of \textit{SmartExchange} with re-training on two compact models with the ImageNet dataset \cite{Deng09imagenet}.}
    \vspace{-0.5em}
    \small{

    \begin{tabular}{p{40pt} p{23pt} p{23pt} p{12pt} p{20pt} p{12pt} p{12pt} p{18pt}} \toprule
    \multirow{2}{*}{Model} & Top-1 & Top-5 & CR & Param. & $B$  & $C_e$ & Spar.  \\
                           & (\%)  & (\%)  &  ($\times$)  &  (MB)  & (MB) & (MB)  & (\%)  \\
    \midrule 
    MBV2 & 72.19\%  & 90.53\%  & - & 13.92 & - & - & - \\
    $\mathrm{MBV2_{SE}}$ & 70.16\%  & 89.54\%  & 6.57 & 2.12 & 0.37 & 1.74 & 0.00 \\
    \midrule
    Eff-B0 & 76.30\% & 93.50\%   & - & 20.40 & - & - & - \\
    $\mathrm{Eff}$-$\mathrm{B0_{SE}}$ & 73.80\%  & 91.79\%  & 6.67 & 3.06 & 0.51 & 2.55 & 0.00 \\
        \bottomrule
    \end{tabular}
    }
    \label{table:compact_results}
     \vspace{-1.5em}
\end{table}
\end{scriptsize}


A more extensive set of evaluation results are summarized in Table~\ref{table:sed_results}, in order to show the maximally achievable gains (and the incurring accuracy losses) by applying \textit{SmartExchange} over the original uncompressed models. In Table~\ref{table:sed_results}, ``CR'' means the \textit{compression rate} in terms of the overall parameter size; ``Param.'', ``$B$'', and ``$C_e$'' denote the total size of the model parameters, the basis matrices, and the coefficient matrices, respectively; ``Spar.'' denotes the ratio of the pruned and total parameters (the higher the better). Without too much surprise, \textit{SmartExchange} compresses the VGG networks by 40$\times$ to 80$\times$, all with negligible (less than 1\%) top-1 accuracy losses. For ResNets, \textit{SmartExchange} is still able to achieve a solid $>$10$\times$ compression ratio. For example, when compressing ResNet50, we find \textit{SmartExchange} to incur almost no accuracy drop, when compressing the model size by 11$\times$ to 14$ \times$.


\textbf{\textit{SmartExchange} applied on compact models. } Table~\ref{table:sed_results} seems to suggest that (naturally) applying \textit{SmartExchange} to more redundant models will have more gains. We thus validate whether the proposed \textit{SmartExchange} algorithm remains to be beneficial, when adopted for well-known compact models, i.e., MobileNetV2 (MBV2) \cite{sandler2018mobilenetv2} and EfficientNet-B0 (Eff-B0) \cite{tan2019efficientnet}. 

As Table \ref{table:compact_results} indicates, despite the original light-weight design, \textit{SmartExchange} still yields promising gains. For example, when compressing MBV2 for $6.57\times$ CR, \textit{SmartExchange} only incurs $\sim$2\% top-1 accuracy and 1\% top-5 accuracy losses. This result is impressive and highly competitive when placed in the context: for example, the latest work \cite{gong2019differentiable} reports $8\times$ compression (4-bit quantization) of MobileNetV2, yet with a 7.07\% top-1 accuracy loss. 

\textbf{Extending \textit{SmartExchange} beyond classification models. } While model compression methods (and hence co-design works) are dominantly evaluated on classification benchmarks, we demonstrate that the effectiveness of \textit{SmartExchange} is beyond one specific task setting. We choose semantic segmentation, a heavily-pursued computer vision task that is well known to be memory/latency/energy-demanding, to apply the proposed algorithm. Specifically, we choose the state-of-the-art DeepLabv3+ \cite{chen2017deeplab} with a ResNet50 backbone (output stride: 16), and the CamVid\cite{brostow2008segmentation} dataset using its standard split. Compared to the original DeepLabv3+, applying \textit{SmartExchange} can lead to 10.86$\times$ CR, with a marginal mean Intersection over Union (mIoU) drop from 74.20\% to 71.20\% (on the validation split). 

\begin{figure}[!t]
    \vspace{-1.5em}
    \centerline{\includegraphics[width=70mm]{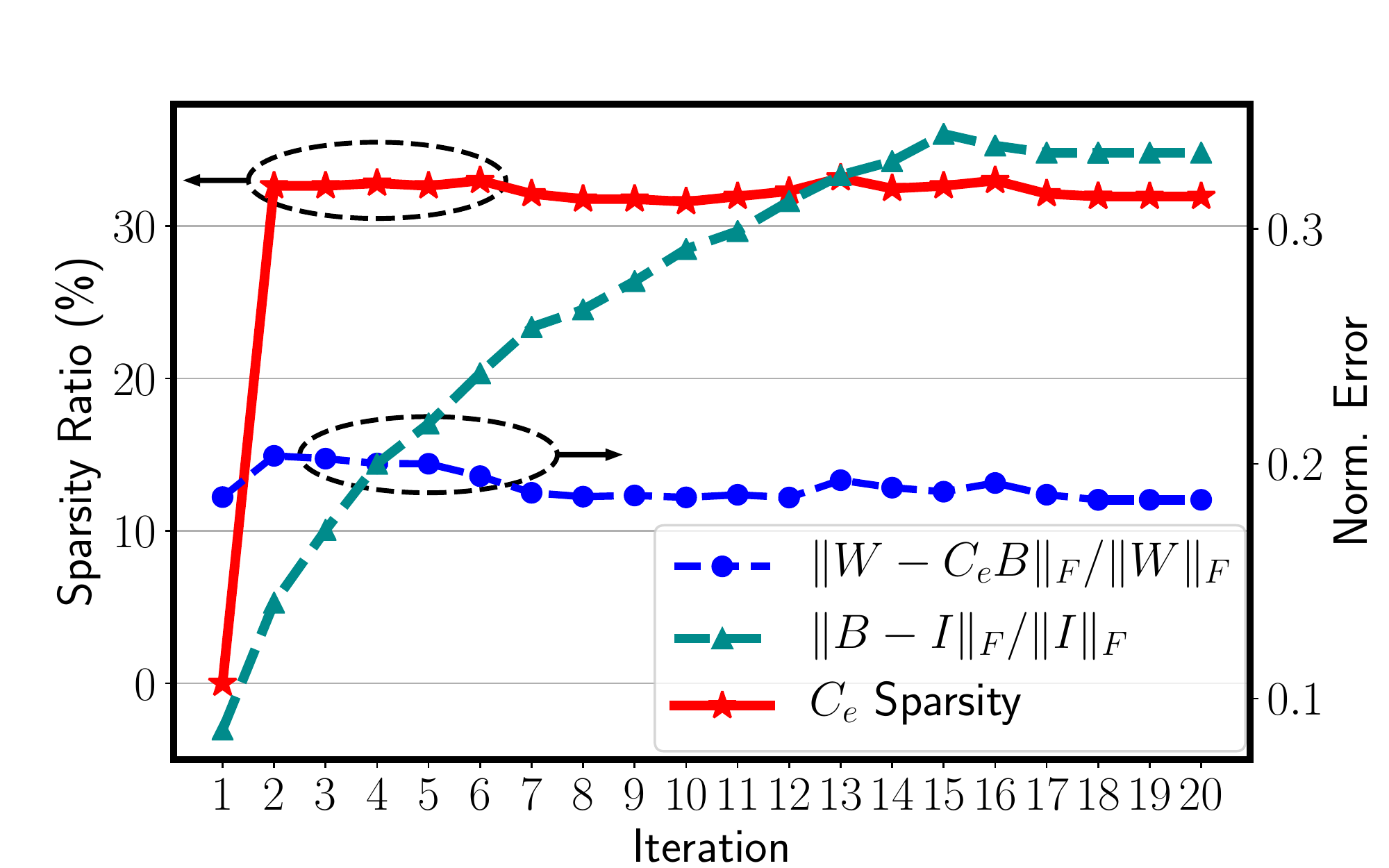}}
    \vspace{-0.7em}
    \caption{{Illustrating an example of the solution evolution during the \textit{SmartExchange} algorithm training.}}
     \vspace{-2em}
    \label{fig:evolution}
\end{figure}

\textbf{\textit{SmartExchange} decomposition evolution. }
To give an example of the decomposition evolution of the \textit{SmartExchange} algorithm, we take one weight matrix $W\in R^{192\times 3}$ from the second CONV layer of the second block in a ResNet164 network pre-trained on CIFAR-10. The \textit{SmartExchange} algorithm decomposes $W=C_e B$, where $C_e\in R^{192\times 3}$ and $B\in R^{3\times 3}$. Figure \ref{fig:evolution} shows the evolution of the reconstruction error, sparsity ratio in $C_e$, and the distance between $B$ and its initialization (identity). We can see that the sparsity ratio in $C_e$ will increase at the beginning at the cost of an increased reconstruction error. But the \textit{SmartExchange} algorithm remedies the error over iterations while maintaining the sparsity. Also, $B$ will gradually become more different from the initialization.


  \vspace{-0.5em}
\subsection{Evaluation of the \textit{SmartExchange} Accelerator.}
  \vspace{-0.3em}
\label{sec:hw-performance}

In this subsection, we present experiments to evaluate the performance of the \textit{SmartExchange} accelerator. Specifically, we first introduce the experiment setup and methodology, and then compare \textit{SmartExchange} accelerator with \textbf{four state-of-the-art DNN accelerators} (covering a diverse range of design considerations) on \textbf{seven DNN models} (including four standard DNNs, two compact models, and one segmentation model) in terms of energy consumption and latency when running on \textbf{three benchmark datasets}. 
Finally, we perform ablation studies for the \textit{SmartExchange} accelerator to quantify and discuss the contribution of its component techniques, its energy breakdown, and its effectiveness in 1) making use of sparsity and 2) dedicated design for handling compact models, aiming to provide more insights.

\begin{table}[!b]
   \vspace{-1.6em}
    \centering
    \caption{The design considerations of the baseline and our accelerators.}
    \vspace{-0.7em}
    \begin{tabular}{@{} c c @{}} 
    \toprule
    Accelerator & Design Considerations \\
    \midrule
     \textbf{DianNao~\cite{chen2014diannao}} & Dense models\\
     \textbf{Cambricon-X~\cite{zhang2016cambricon}} & Unstructure weight sparsity \\
     \multirow{2}{*}{\textbf{SCNN~\cite{parashar2017scnn}}} & Unstructure weight sparsity \\
                                                            &  + Activation sparsity \\
     \textbf{Bit-pragmatic~\cite{albericio2017bit}} & Bit-level activation sparsity \\
     \midrule
     \multirow{2}{*}{\textbf{Ours}} & Vector-wise weight sparsity \\
                                    &  + Bit-level and vector-wise activation sparsity \\
    \bottomrule
    \end{tabular}
    \label{table:considerations}
\end{table}

\begin{table}[!t]
    \centering
    \caption{A summary of the computation and storage resources in the \textit{SmartExchange} and baseline accelerators.}
    \vspace{-0.8em}
    \begin{tabular}{@{}r l r l @{}} 
    \toprule
    \multicolumn{4}{c}{\textbf{\textit{SmartExchange} and Bit-pragmatic~\cite{albericio2017bit} }} \\
    \midrule 
     \textbf{$dim_M$} & 64 & \textbf{Input GB} & 16KB$\times$32Banks   \\
     \textbf{$dim_C$} & 16 & \textbf{Output GB} & 2KB$\times$2Banks    \\
     \textbf{$dim_F$} & 8  & \textbf{Weight Buff./slice} &  2KB$\times$2Banks \\
     \textbf{\# of bit-serial mul.} & 8K  & \textbf{Precision} &  8 bits \\
    \midrule 
    \multicolumn{4}{c}{\textbf{DianNao~\cite{chen2014diannao}, SCNN~\cite{parashar2017scnn}, and Cambricon-X~\cite{zhang2016cambricon}  }} \\
    \midrule 
     \multicolumn{4}{l}{The same total on-chip SRAM storage as \textit{SmartExchange}} \\
     \textbf{\# of 8-bit mul.} & 1K  & \textbf{Precision} &  8 bits \\
    \bottomrule
    \end{tabular}
    \label{table:baselines}
     \vspace{-1.8em}
\end{table}

\textbf{Experiment setup and methodology.} 
\ul{Baselines and configurations:} we benchmark the \textit{SmartExchange} accelerator with
four state-of-the-art accelerators: DianNao~\cite{chen2014diannao}, SCNN~\cite{parashar2017scnn}, Cambricon-X~\cite{zhang2016cambricon}, and Bit-pragmatic~\cite{albericio2017bit}. These representative accelerators have demonstrated promising acceleration performance, and are designed with a diverse design considerations as summarized in Table~\ref{table:considerations}. Specifically, DianNao~\cite{chen2014diannao} is a classical architecture for DNN inference which is reported to be over 100$\times$ faster and over 20$\times$ more energy efficient than those of CPUs. While DianNao considers dense models, the other three accelerators take advantage of certain kinds of sparsity in DNNs. To ensure fair comparisons, we assign the \textit{SmartExchange} accelerator and baselines with the same computation resources and on-chip SRAM storage in all experiments, as listed in Table~\ref{table:baselines}. For example, the DianNao, SCNN and Cambricon-X accelerators use 1K 8-bit non-bit-serial multipliers and 
\textit{SmartExchange} and Bit-pragmatic employ an equivalent 8K bit-serial multipliers.

\begin{figure} [!t]
    \centering
    \includegraphics[width=\linewidth]{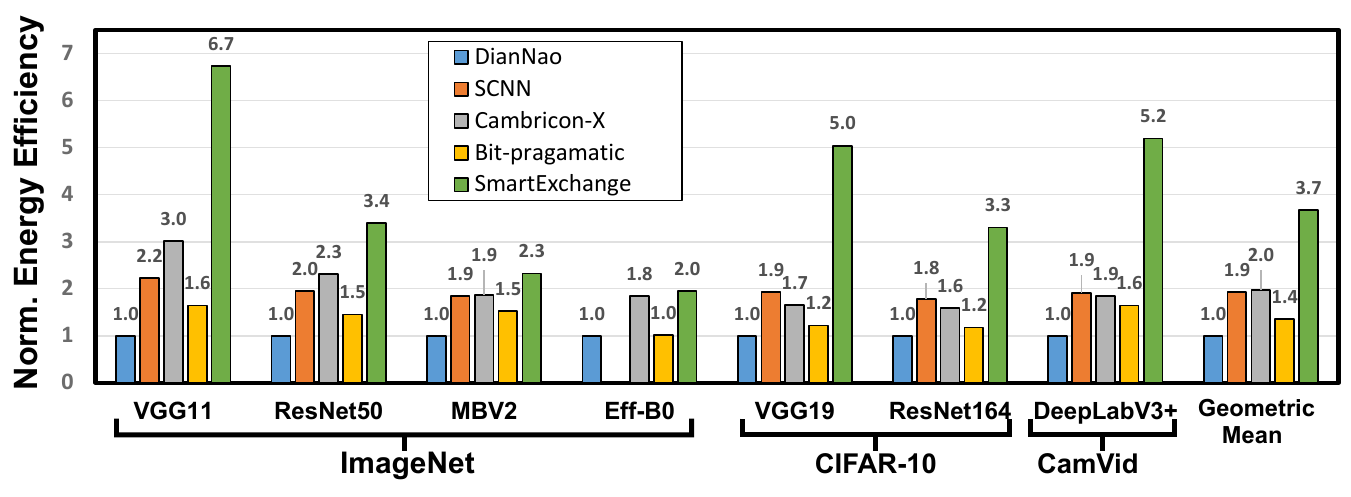}
    \vspace{-2em}
    \caption{The normalized energy efficiency (over DianNao) achieved by the \textit{SmartExchange} accelerator over the four state-of-the-art baseline accelerators on seven DNN models and three datasets.}
    \label{fig:energy_all}
     \vspace{-1.8em}
\end{figure}

\begin{figure} [!t]
    \centering
    \includegraphics[width=\linewidth]{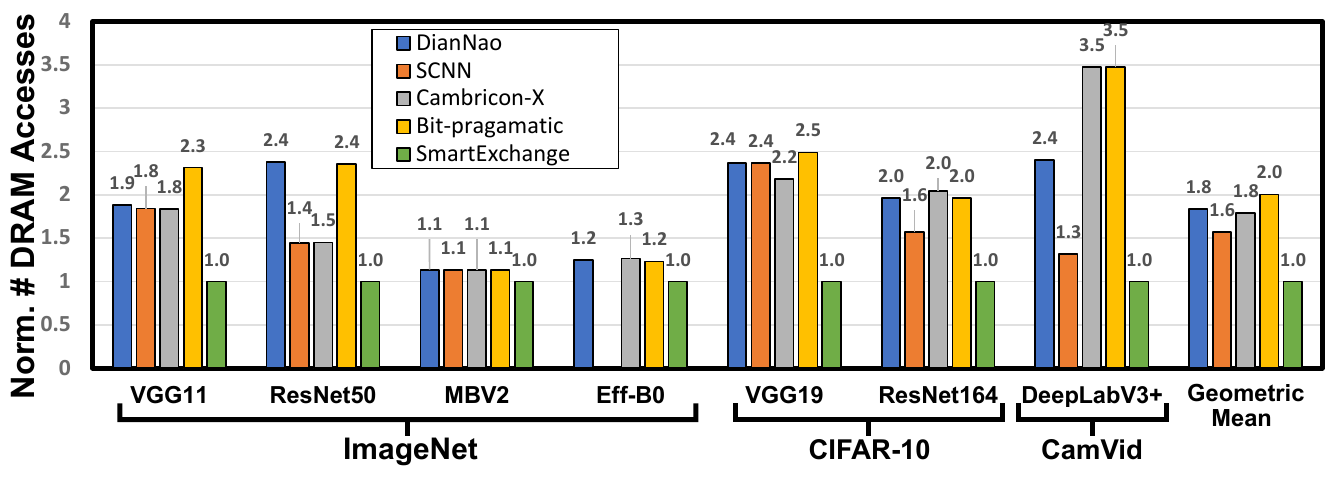}
     \vspace{-2em}
    \caption{The normalized number of DRAM accesses (over the \textit{SmartExchange} accelerator) of the  \textit{SmartExchange} and four state-of-the-art baseline accelerators on seven DNN models and three datasets.}
    \label{fig:dram}
     \vspace{-2em}
\end{figure}

\begin{figure} [!b]
    \centering
  \vspace{-1.8em}
    \includegraphics[width=\linewidth]{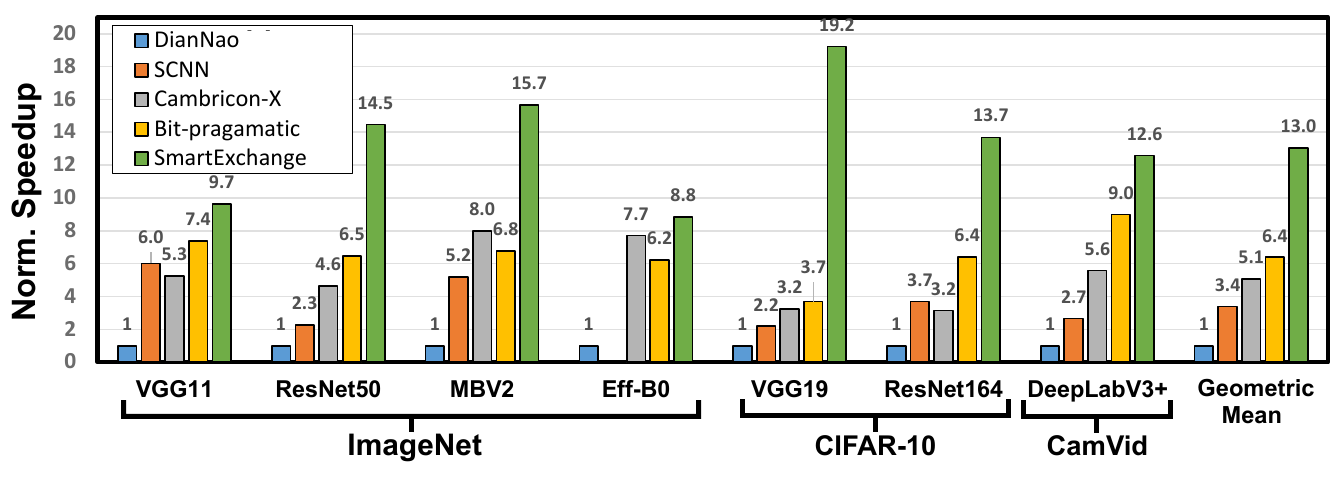}
    \vspace{-2em}
    \caption{The normalized speedup (over DianNao) achieved by the \textit{SmartExchange} accelerator over the four state-of-the-art baseline accelerators on seven DNN models and three datasets.}
    \label{fig:speedup_all}
\end{figure}

\begin{figure} [!t]
    \centering   
    \includegraphics[width=\linewidth]{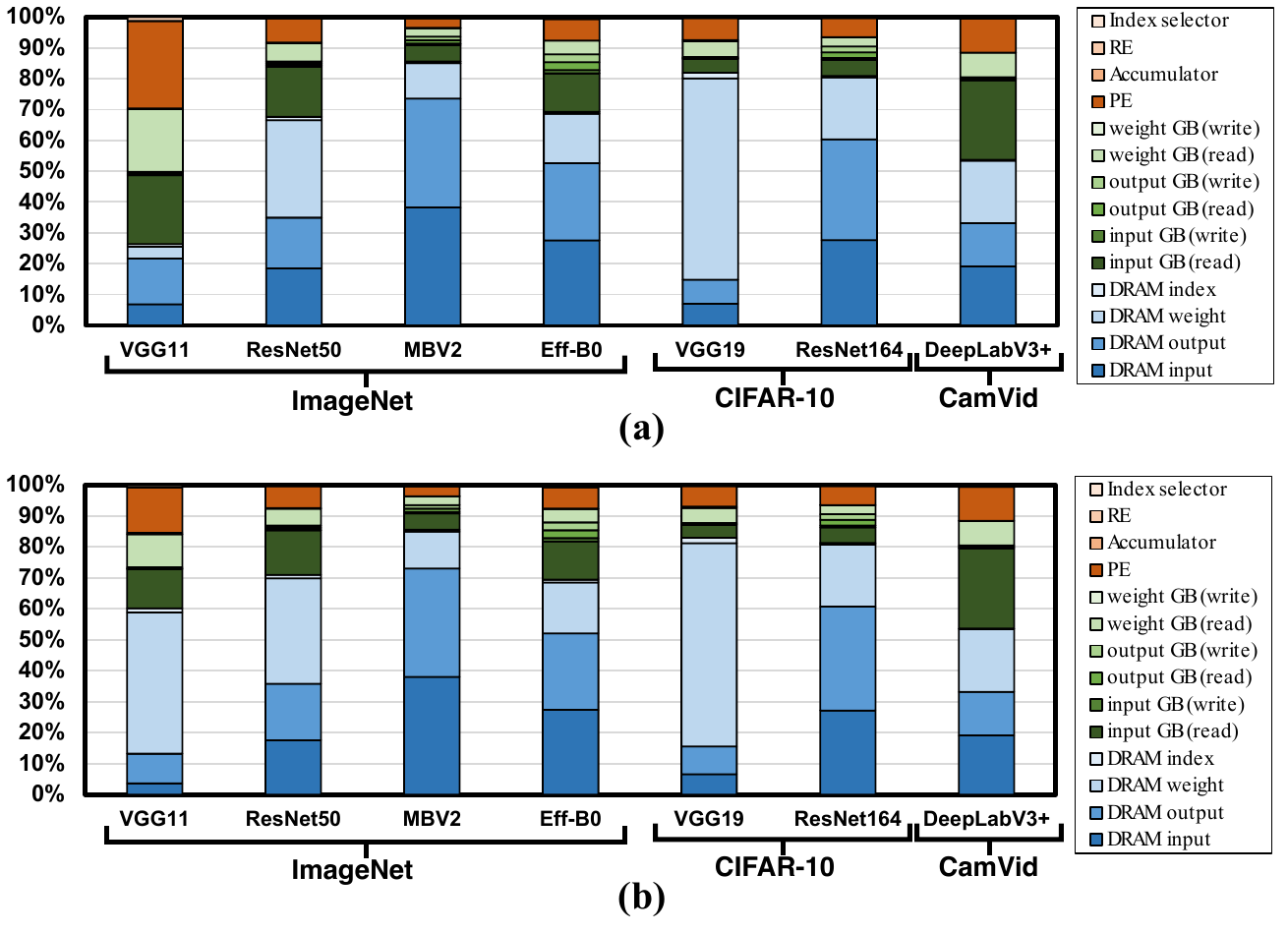}
    \vspace{-1.8em}
    \caption{The energy breakdown of the \textit{SmartExchange} accelerator when running the (a) CONV and squeeze-and-excite layers and (b) CONV, squeeze-and-excite, and FC layers (all types of layers) of seven DNN models on three datasets.}
    \vspace{-1.5em}
    \label{fig:energy_breakdown}
\end{figure}

\begin{figure} [!t]
    \centering   
    \includegraphics[width=\linewidth]{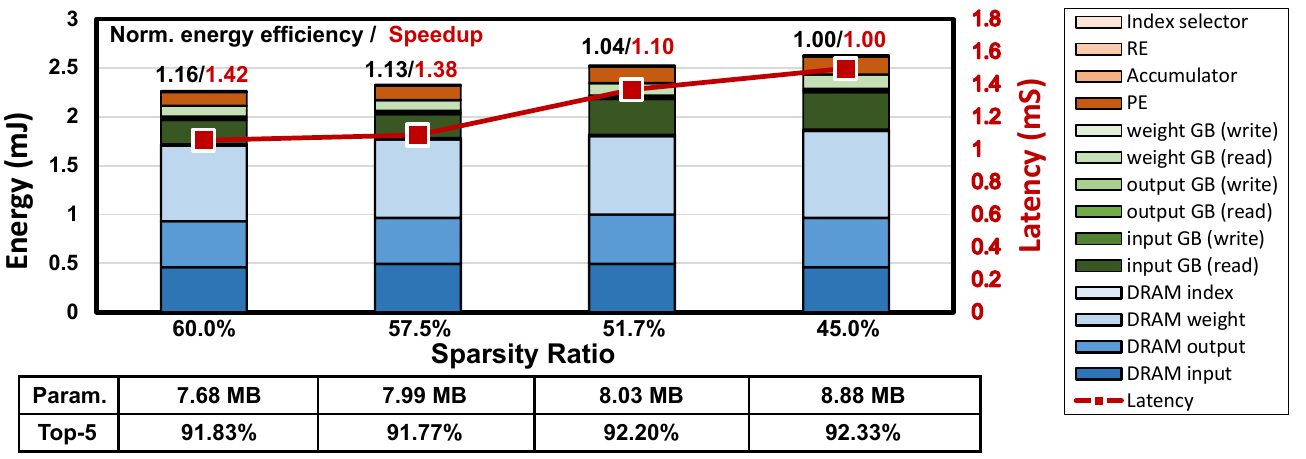}
     \vspace{-1.6em}
    \caption{The energy breakdown and latency of the \textit{SmartExchange} accelerator when running ResNet50 with four different sparsity ratios.}
    \label{fig:dff_ResNet50}
    \vspace{-2em}
\end{figure}

For handling the dynamic sparsity in the \textit{SmartExchange} accelerator, the on-chip input GB bandwidth and weight GB bandwidth with each PE slice are set to be four and two times of those in the corresponding dense models, respectively, which are empirically found to be sufficient for handling all the considered models and datasets. Meanwhile, because the computation resources for the baseline accelerators may be different from their original papers, the bandwidth settings are configured accordingly based on their papers' reported design principles. Note that 1) we do not consider FC layers when benchmarking the \textit{SmartExchange} accelerator with the baseline accelerators (see \Cref{fig:energy_all,fig:dram,fig:speedup_all}) for a fair comparison as the SCNN~\cite{parashar2017scnn} baseline is designed for CONV layers, and similarly, we do not consider EfficientNet-B0 for the SCNN accelerator as SCNN is not designed for handling the squeeze-and-excite layers adopted in EfficientNet-B0; 2) our ablation studies consider all layers in the models (see \Cref{fig:energy_breakdown,fig:dff_ResNet50}).


\ul{Benchmark models, datasets, and precision:}
We use seven representative DNNs (ResNet50, ResNet164, VGG11, VGG19, MobileNetV2, EfficientNet-B0, and DeepLabV3+) and three benchmark datasets (CIFAR-10~\cite{cifar10}, ImageNet~\cite{Deng09imagenet}, and CamVid~\cite{brostow2008segmentation}). Regarding the precision, we adopt 1) 8-bit activations for both the baseline-used and \textit{SmartExchange}-based DNNs; and 2) 8-bit weights in the baseline-used DNNs, and 8-bit/4-bit precision for the basis and coefficient matrices in the \textit{SmartExchange}-based DNNs.

\ul{Technology-dependent parameters:} 
For evaluating the performance of the \textit{SmartExchange} accelerator, we implemented a custom cycle-accurate simulator, aiming to model the Register-Transfer-Level (RTL) behavior of synthesized circuits, and verified the simulator against the corresponding RTL implementation to ensure its correctness. Specifically, 
the gate-level netlist and SRAM are generated based on a commercial 28nm technology using the Synopsys Design Compiler and Arm Artisan Memory Compilers, proper activity factors are set at the input ports of the memory/computation units, and the energy is calculated using a state-of-the-art tool PrimeTime PX~\cite{PTPX}. 
Meanwhile, thanks to the clear description of the baseline accelerators' papers and easy representation of their works, we followed their designs and implemented custom cycle-accurate simulators for all the baselines. In this way, we can evaluate the performance of both the baseline and our accelerators based on the same commercial 28nm technology. 
The resulting designs operate at a frequency of 1GHz and the performance results are normalized over that of the DianNao accelerator, where the DianNao design is modified to ensure that all accelerators have the same hardware resources (see Table \ref{table:baselines}). 
We refer to ~\cite{yang2018dnn} for the unit energy of DRAM accesses, which is 100pJ per 8 bit, and the unit energy costs for computation and SRAM accesses are listed in Table \ref{table:unit_energy}.

\textbf{\textit{SmartExchange} vs. state-of-the-art accelerators.} 
\ul{Energy efficiency over that of the baseline accelerators:}
Figure~\ref{fig:energy_all} shows the normalized energy efficiency of the \textit{SmartExchange} and the baseline accelerators. It is shown that the \textit{SmartExchange} accelerator consumes the least energy under all the considered DNN models and datasets, achieving an energy efficiency improvement ranging from $2.0\times$ to $6.7\times$. The \textit{SmartExchange} accelerator's outstanding energy efficiency performance is a result of \textit{SmartExchange}'s algorithm-hardware co-design effort to effectively trade the much higher-cost memory storage/accesses for the lower-cost computations (i.e., rebuilding the weights using the basis and coefficient matrices at the least costly RF and PE levels vs. fetching them from the DRAM). Note that \textit{SmartExchange} non-trivially outperforms all baseline accelerators even on the compact models (i.e., MobileNetV2 and EfficientNet-B0) thanks to both the \textit{SmartExchange} algorithm's higher compression ratio and the \textit{SmartExchange} accelerator's dedicated and effective design (see Section \ref{subsec:accelerator}) of handling depth-wise CONV and squeeze-and-excite layers that are commonly adopted in compact models.



Figure~\ref{fig:dram} shows the normalized number of DRAM accesses for the weights and input/output activations. We can see that: 
1) the baselines always require more (1.1$\times$ to 3.5$\times$) DRAM accesses than the \textit{SmartExchange} accelerator, e.g.,  see the ResNet and VGG models on the ImageNet and CIFAR-10 datasets as well as the segmentation model DeepLabV3+ on the CamVid dataset;
2) \textit{SmartExchange}’s DRAM-access reduction is smaller when the models' activations dominate the cost (e.g., compact DNN models);
and 3) the \textit{SmartExchange} accelerator can reduce the number of DRAM accesses over the baselines by up to 1.3$\times$ for EfficientNet-B0, indicating the effectiveness of our dedicated design for handling the squeeze-and-excite layers (see Section \ref{subsec:accelerator}).

\ul{Speedup over that of the baseline accelerators:} Similar to benchmarking the \textit{SmartExchange} accelerator's energy efficiency, we compare its latency of processing one image (i.e., batch size is 1) over that of the baseline accelerators on various DNN models and datasets, as shown in Figure~\ref{fig:speedup_all}. We can see that the \textit{SmartExchange} accelerator achieves the best performance under all the considered DNN models and datasets, achieving a latency improvement ranging from $8.8\times$ to $19.2\times$. Again, this experiment validates the effectiveness of
\textit{SmartExchange}'s algorithm-hardware co-design effort to reduce the latency on fetching both the weights and the activations from the memories to the computation resources. Since the \textit{SmartExchange} accelerator takes advantage of both the weights' vector-wise sparsity and the activations' bit-level and vector-wise sparsity, it has a higher speedup over all the baselines that make use of only one kind of sparsity. Specifically, the \textit{SmartExchange} accelerator has an average latency improvement of 3.8$\times$, 2.5$\times$, and 2.0$\times$ over SCNN~\cite{parashar2017scnn} and Cambricon-X~\cite{zhang2016cambricon} which consider unstructured sparsity, and Bit-pragmatic~\cite{albericio2017bit} which considers the bit-level sparsity in activations, respectively. 

\textbf{Contributions of \textit{SmartExchange}'s component techniques.}
The aforementioned energy efficiency and latency improvement of the \textit{SmartExchange} accelerator comes from the algorithm-hardware co-design efforts including the \textit{SmartExchange} algorithm's model compression (see Section \ref{sec:algorithm}) and the \textit{SmartExchange} accelerator's support for both vector-wise sparsity (i.e., index selecting) and bit-level sparsity (i.e., bit-serial multiplier) (see Section \ref{subsec:accelerator}).
To quantify the contribution of \textit{SmartExchange}'s component techniques, we build a similar baseline  accelerator as the \textit{SmartExchange} accelerator and run a dense DNN on the baseline accelerator. Specifically, the baseline accelerator uses non-bit-serial multipliers, $dim_M$=16, $dim_C$=8, and $dim_F$=8 to ensure the required hardware resources to be the same as that of the \textit{SmartExchange} accelerator.
When running ResNet50, the \textit{SmartExchange} accelerator achieves 3.65$\times$ better energy efficiency than the baseline accelerator, where the reduced DRAM accesses resulted from the \textit{SmartExchange}'s model compression, vector-wise sparsity support, and bit-level sparsity support contribute to 23.99\%, 12.48\%, and 36.14\% of the total energy savings, respectively.
Assuming a sufficient DRAM bandwidth, the \textit{SmartExchange} accelerator achieves 7.41$\times$ speedup than the baseline accelerator, thanks to its 1) effort to leverage the sparsity to reduce unnecessary data movements and computations and 2) increased parallel computation resources (note that the number of bit-serial multipliers is 8$\times$ of that of non-bit-serial multipliers given the same computation resource).


\textbf{The \textit{SmartExchange} accelerator's energy breakdown.}
Figure~\ref{fig:energy_breakdown} (a) shows the \textit{SmartExchange} accelerator's energy breakdown in terms of computations and accessing various memory hierarchies, when processing only the CONV and squeeze-and-excite layers (i.e., excluding the FC layers) of various DNN models and datasets. We can see that 1) the energy cost of accessing DRAM is dominated by the input/output activations for most of the models (i.e., see the VGG11, MobileNetV2, and EfficientNet-B0 models on the ImageNet dataset, the ResNet164 on the CIFAR-10 dataset, and the DeepLabV3+ model on the CamVid dataset), because the \textit{SmartExchange} algorithm can largely reduce the number of weight accesses from the DRAM; 2) the energy cost of accessing DRAM for the weights is still dominant in models where the model sizes are very large, e.g., see the VGG19 model on the CIFAR10 dataset and the ResNet50 model on the ImageNet dataset; and 3) the RE and index selector only account for $<$0.78\% and $<$0.05\% of the total energy cost, which are negligible.

When considering all layers (see Figure~\ref{fig:energy_breakdown} (b)), the trends of the experiment results are similar to those in Figure~\ref{fig:energy_breakdown} (a), except for the VGG11 model. This is because the FC layers in most of the models consume only $<$7.77\% of the total energy cost, whereas the FC weight DRAM accesses in VGG11 account for up to 43.08\% of the total energy cost and up to 95.66\% of the total parameter size.
Note that although the total size of the \textit{SmartExchange}-compressed weights is similar for the VGG19 and ResNet164 models on CIFAR10 (see Table~\ref{table:sed_results}), their weight DRAM accesses cost percentages are very different. This is because 1) the original ResNet164 model has much more activations than that of the VGG19 model and 2) the activations in the VGG19 model \cite{luo2017thinet} have been largely pruned thanks to the models' high filter-wise sparsity (e.g., 90.79\%) which enables pruning the whole filters and their corresponding activations (e.g., enabling 81.04\% and 26.64\% of the input and output activations to be pruned), both leading to the large gap in the cost percentage of the weight DRAM accesses in the two models.

\textbf{\textit{SmartExchange}'s effectiveness in exploiting sparsity.} 
Figure~\ref{fig:dff_ResNet50} shows the normalized energy consumption and latency (over the total energy cost and latency of the models) when the \textit{SmartExchange} accelerator processes ResNet50 with four vector-wise weight sparsity ratios, where the corresponding model size and accuracy are summarized in the bottom-left corner. We can see that: 1) the total energy cost of the input activations' DRAM and GB accesses is reduced by 18.33\% when the weight sparsity increases by 15\% (from 45.0\% to  60.0\%), showing that our accelerator can effectively utilize the vector-wise weight sparsity to save the energy cost of accessing both the sparse weights and the corresponding inputs; and 2) the latency is reduced by 41.83\% when the weight sparsity increases from 45.0\% to  60.0\%, indicating the \textit{SmartExchange} accelerator can indeed utilize the vector-wise weight sparsity to skip the corresponding input accesses and computations to reduce latency.

\begin{figure} [!t]
    \centering
    \includegraphics[width=\linewidth]{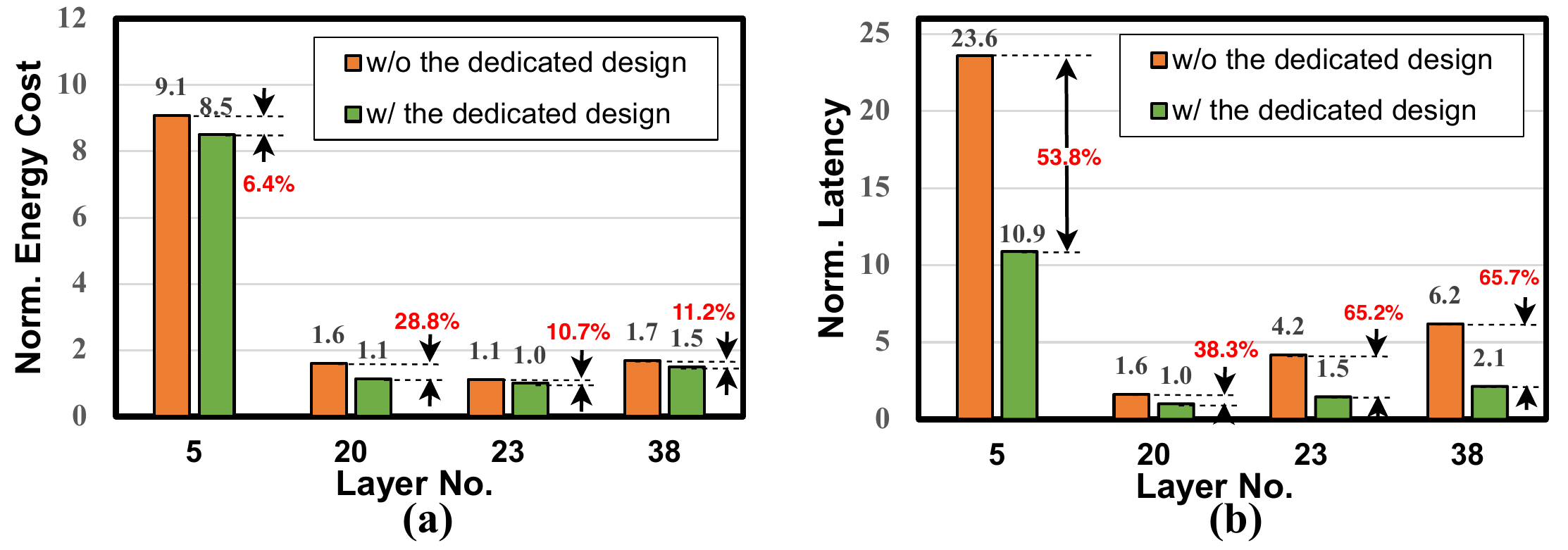}
    \vspace{-2em}
    \caption{The normalized (a) energy cost and (b) latency of the depth-wise CONV layers w/ and w/o the proposed dedicated design for compact models, when processing MobileNetV2 on the ImageNet dataset. }
    \label{fig:depth_ablation}
    \vspace{-1.6em}
\end{figure}

\textbf{Effectiveness of \textit{SmartExchange}'s support for compact models.}
We perform an ablation experiment to evaluate the \textit{SmartExchange} accelerator's dedicated design including optimized dataflow and PE line configuration (see Section \ref{subsec:accelerator}) for handling compact models.
Figure \ref{fig:depth_ablation} (a) shows the normalized layer-wise energy cost on selected depth-wise CONV layers of MobileNetV2 with and without the proposed dedicated design. We can see that the proposed design can effectively reduce the energy cost by up to 28.8\%. Meanwhile, Figure \ref{fig:depth_ablation} (b) further shows that the normalized layer-wise latency can be reduced by 38.3\% to 65.7\%.

 \vspace{-0.5em}
\section{Related Works}\label{sec:related_work}
\textbf{Compression-aware DNN accelerators.} 
To achieve aggressive performance improvement, researchers have explored from both the algorithm and architecture sides. In general, there exist three typical algorithm approaches, weight \textit{decomposition}, data \textit{quantization}, and weight \textit{sparsification}, that have been exploited by hardware design. H. Huang et al. \cite{huang2018highly} demonstrate DNNs with tensorized decomposition on non-volatile memory (NVM) devices. 
For the weight sparsification accelerators, \cite{han2016eie, parashar2017scnn, zhang2016cambricon} have been proposed for making use of unstructured sparsity. Cambricon-S \cite{zhou2018cambricon} proposes a co-designed weight sparsity pattern to reduce irregularity.  
Most of recent accelerators use equal or less than 16-bit fixed-point quantized data \cite{parashar2017scnn, zhang2016cambricon}. The works in \cite{han2016eie, zhou2018cambricon} uses clustering to further encode weights; Stripes \cite{judd2016stripes} and UNPU \cite{lee2018unpu} leverage a bit-serial processing to support flexible bit widths to better balance the accuracy loss and performance improvement; Bit-pragmatic \cite{albericio2017bit} utilizes the input bit-level sparsity to improve throughput and energy efficiency; and Bit-Tactical \cite{delmas2019bit} combines the weight unstructured sparsity with input bit-level sparsity.
To our best knowledge, \textit{SmartExchange} is the first formulation that unifies weight \textit{decomposition}, \textit{quantization}, and \textit{sparsification} (especially vector-wise structured sparsity) approaches to simultaneously shrink the memory footprint and simplify the computations when recovering the weight matrix during runtime.

\vspace{-0.3em}
\section{Conclusion}\label{sec:conclusion}
We propose \textit{SmartExchange}, an algorithm-hardware co-design framework to trade higher-cost memory storage/access for lower-cost computation, for boosting the energy efficiency and speed of DNN inference. 
Extensive experiments show that the \textit{SmartExchange} algorithm outperforms state-of-the-art compression techniques on seven DNN models and three datasets under various settings, while the \textit{SmartExchange} accelerator outperforms state-of-the-art DNN accelerators in terms of both energy efficiency and latency (up to 6.7$\times$ and 19.2$\times$, respectively).

\vspace{-0.5em}
\section*{Acknowledgment}
\vspace{-0.5em}
We thank Dr. Lei Deng (UCSB) for his comments and suggestions. 
This work is supported in part by the NSF RTML grant 1937592, 1937588, and NSF 1838873, 1816833, 1719160, 1725447, 1730309.



\bibliographystyle{IEEEtranS}

\begin{thebibliography}{10}
\providecommand{\url}[1]{#1}
\csname url@samestyle\endcsname
\providecommand{\newblock}{\relax}
\providecommand{\bibinfo}[2]{#2}
\providecommand{\BIBentrySTDinterwordspacing}{\spaceskip=0pt\relax}
\providecommand{\BIBentryALTinterwordstretchfactor}{4}
\providecommand{\BIBentryALTinterwordspacing}{\spaceskip=\fontdimen2\font plus
\BIBentryALTinterwordstretchfactor\fontdimen3\font minus
  \fontdimen4\font\relax}
\providecommand{\BIBforeignlanguage}[2]{{%
\expandafter\ifx\csname l@#1\endcsname\relax
\typeout{** WARNING: IEEEtranS.bst: No hyphenation pattern has been}%
\typeout{** loaded for the language `#1'. Using the pattern for}%
\typeout{** the default language instead.}%
\else
\language=\csname l@#1\endcsname
\fi
#2}}
\providecommand{\BIBdecl}{\relax}
\BIBdecl

\bibitem{albericio2017bit}
J.~Albericio, A.~Delm{\'a}s, P.~Judd, S.~Sharify, G.~O'Leary, R.~Genov, and
  A.~Moshovos, ``Bit-pragmatic deep neural network computing,'' in
  \emph{Proceedings of the 50th Annual IEEE/ACM International Symposium on
  Microarchitecture}.\hskip 1em plus 0.5em minus 0.4em\relax ACM, 2017, pp.
  382--394.

\bibitem{Boris_ISSCC}
D.~{Bankman}, L.~{Yang}, B.~{Moons}, M.~{Verhelst}, and B.~{Murmann}, ``An
  always-on 3.8μj/86\% cifar-10 mixed-signal binary cnn processor with all
  memory on chip in 28nm cmos,'' in \emph{2018 IEEE International Solid - State
  Circuits Conference - (ISSCC)}, Feb 2018, pp. 222--224.

\bibitem{banner2018scalable}
R.~Banner, I.~Hubara, E.~Hoffer, and D.~Soudry, ``Scalable methods for 8-bit
  training of neural networks,'' in \emph{Advances in neural information
  processing systems}, 2018, pp. 5145--5153.

\bibitem{brostow2008segmentation}
G.~J. Brostow, J.~Shotton, J.~Fauqueur, and R.~Cipolla, ``Segmentation and
  recognition using structure from motion point clouds,'' in \emph{European
  conference on computer vision}.\hskip 1em plus 0.5em minus 0.4em\relax
  Springer, 2008, pp. 44--57.

\bibitem{chen2017deeplab}
L.-C. Chen, G.~Papandreou, I.~Kokkinos, K.~Murphy, and A.~L. Yuille, ``Deeplab:
  Semantic image segmentation with deep convolutional nets, atrous convolution,
  and fully connected crfs,'' \emph{IEEE transactions on pattern analysis and
  machine intelligence}, vol.~40, no.~4, pp. 834--848, 2017.

\bibitem{chen2014diannao}
T.~Chen, Z.~Du, N.~Sun, J.~Wang, C.~Wu, Y.~Chen, and O.~Temam, ``Diannao: A
  small-footprint high-throughput accelerator for ubiquitous
  machine-learning,'' in \emph{Proceedings of the 19th International Conference
  on Architectural Support for Programming Languages and Operating Systems},
  ser. ASPLOS ’14, 2014, p. 269–284.

\bibitem{eyeriss}
Y.-H. Chen, J.~Emer, and V.~Sze, ``Eyeriss: A spatial architecture for
  energy-efficient dataflow for convolutional neural networks,'' in
  \emph{Computer Architecture (ISCA), 2016 ACM/IEEE 43th Annual International
  Symposium on}.\hskip 1em plus 0.5em minus 0.4em\relax IEEE Press, 2016, pp.
  367--379.

\bibitem{chen2017eyeriss}
Y.-H. Chen, T.~Krishna, J.~S. Emer, and V.~Sze, ``Eyeriss: An energy-efficient
  reconfigurable accelerator for deep convolutional neural networks,''
  \emph{IEEE Journal of Solid-State Circuits}, vol.~52, no.~1, pp. 127--138,
  2017.

\bibitem{chen2019eyeriss}
Y.-H. Chen, T.-J. Yang, J.~Emer, and V.~Sze, ``Eyeriss v2: A flexible
  accelerator for emerging deep neural networks on mobile devices,'' \emph{IEEE
  Journal on Emerging and Selected Topics in Circuits and Systems}, 2019.

\bibitem{delmas2019bit}
A.~Delmas~Lascorz, P.~Judd, D.~M. Stuart, Z.~Poulos, M.~Mahmoud, S.~Sharify,
  M.~Nikolic, K.~Siu, and A.~Moshovos, ``Bit-tactical: A software/hardware
  approach to exploiting value and bit sparsity in neural networks,'' in
  \emph{Proceedings of the Twenty-Fourth International Conference on
  Architectural Support for Programming Languages and Operating Systems}.\hskip
  1em plus 0.5em minus 0.4em\relax ACM, 2019, pp. 749--763.

\bibitem{Deng09imagenet}
J.~Deng, W.~Dong, R.~Socher, L.~jia Li, K.~Li, and L.~Fei-fei, ``Imagenet: A
  large-scale hierarchical image database,'' in \emph{Proceedings of the IEEE
  conference on computer vision and pattern recognition}, 2009.

\bibitem{denton2014exploiting}
E.~L. Denton, W.~Zaremba, J.~Bruna, Y.~LeCun, and R.~Fergus, ``Exploiting
  linear structure within convolutional networks for efficient evaluation,'' in
  \emph{Advances in neural information processing systems}, 2014, pp.
  1269--1277.

\bibitem{du2015shidiannao}
Z.~Du, R.~Fasthuber, T.~Chen, P.~Ienne, L.~Li, T.~Luo, X.~Feng, Y.~Chen, and
  O.~Temam, ``Shidiannao: Shifting vision processing closer to the sensor,'' in
  \emph{2015 ACM/IEEE 42nd Annual International Symposium on Computer
  Architecture (ISCA)}, vol.~43, no.~3.\hskip 1em plus 0.5em minus 0.4em\relax
  ACM, 2015, pp. 92--104.

\bibitem{gong2019differentiable}
R.~Gong, X.~Liu, S.~Jiang, T.~Li, P.~Hu, J.~Lin, F.~Yu, and J.~Yan,
  ``Differentiable soft quantization: Bridging full-precision and low-bit
  neural networks,'' in \emph{Proceedings of the IEEE International Conference
  on Computer Vision}, 2019, pp. 4852--4861.

\bibitem{gong2014compressing}
Y.~Gong, L.~Liu, M.~Yang, and L.~Bourdev, ``Compressing deep convolutional
  networks using vector quantization,'' \emph{arXiv preprint arXiv:1412.6115},
  2014.

\bibitem{gui2019adversarially}
S.~Gui, H.~Wang, C.~Yu, H.~Yang, Z.~Wang, and J.~Liu, ``Adversarially trained
  model compression: When robustness meets efficiency,'' 2019.

\bibitem{gui2019model}
S.~Gui, H.~N. Wang, H.~Yang, C.~Yu, Z.~Wang, and J.~Liu, ``Model compression
  with adversarial robustness: A unified optimization framework,'' in
  \emph{Advances in Neural Information Processing Systems}, 2019, pp.
  1283--1294.

\bibitem{han2016eie}
S.~Han, X.~Liu, H.~Mao, J.~Pu, A.~Pedram, M.~A. Horowitz, and W.~J. Dally,
  ``Eie: efficient inference engine on compressed deep neural network,'' in
  \emph{2016 ACM/IEEE 43rd Annual International Symposium on Computer
  Architecture (ISCA)}.\hskip 1em plus 0.5em minus 0.4em\relax IEEE, 2016, pp.
  243--254.

\bibitem{han2015deep}
S.~Han, H.~Mao, and W.~J. Dally, ``Deep compression: Compressing deep neural
  networks with pruning, trained quantization and huffman coding,''
  \emph{International Conference on Learning Representations}, 2016.

\bibitem{han2015learning}
S.~Han, J.~Pool, J.~Tran, and W.~Dally, ``Learning both weights and connections
  for efficient neural network,'' in \emph{Advances in neural information
  processing systems}, 2015, pp. 1135--1143.

\bibitem{he2017channel}
Y.~He, X.~Zhang, and J.~Sun, ``Channel pruning for accelerating very deep
  neural networks,'' in \emph{Proceedings of the IEEE International Conference
  on Computer Vision (ICCV)}, 2017, pp. 1389--1397.

\bibitem{howard2017mobilenets}
A.~G. Howard, M.~Zhu, B.~Chen, D.~Kalenichenko, W.~Wang, T.~Weyand,
  M.~Andreetto, and H.~Adam, ``Mobilenets: Efficient convolutional neural
  networks for mobile vision applications,'' \emph{arXiv preprint
  arXiv:1704.04861}, 2017.

\bibitem{huang2018highly}
H.~Huang, L.~Ni, K.~Wang, Y.~Wang, and H.~Yu, ``A highly parallel and energy
  efficient three-dimensional multilayer cmos-rram accelerator for tensorized
  neural network,'' \emph{IEEE Transactions on Nanotechnology}, vol.~17, no.~4,
  pp. 645--656, 2018.

\bibitem{jin2014flattened}
J.~Jin, A.~Dundar, and E.~Culurciello, ``Flattened convolutional neural
  networks for feedforward acceleration,'' \emph{arXiv preprint
  arXiv:1412.5474}, 2014.

\bibitem{judd2016stripes}
P.~Judd, J.~Albericio, T.~Hetherington, T.~M. Aamodt, and A.~Moshovos,
  ``Stripes: Bit-serial deep neural network computing,'' in \emph{2016 49th
  Annual IEEE/ACM International Symposium on Microarchitecture (MICRO)}.\hskip
  1em plus 0.5em minus 0.4em\relax IEEE, 2016, pp. 1--12.

\bibitem{cifar10}
\BIBentryALTinterwordspacing
A.~Krizhevsky, V.~Nair, and G.~Hinton, ``Cifar-10 (canadian institute for
  advanced research).'' [Online]. Available:
  \url{http://www.cs.toronto.edu/~kriz/cifar.html}
\BIBentrySTDinterwordspacing

\bibitem{lee2018unpu}
J.~Lee, C.~Kim, S.~Kang, D.~Shin, S.~Kim, and H.-J. Yoo, ``Unpu: A 50.6 tops/w
  unified deep neural network accelerator with 1b-to-16b fully-variable weight
  bit-precision,'' in \emph{2018 IEEE International Solid-State Circuits
  Conference-(ISSCC)}.\hskip 1em plus 0.5em minus 0.4em\relax IEEE, 2018, pp.
  218--220.

\bibitem{li2016pruning}
H.~Li, A.~Kadav, I.~Durdanovic, H.~Samet, and H.~P. Graf, ``Pruning filters for
  efficient convnets,'' \emph{International Conference on Learning
  Representations}, 2017.

\bibitem{Ziyun-ISSC2019}
Z.~{Li}, Y.~{Chen}, L.~{Gong}, L.~{Liu}, D.~{Sylvester}, D.~{Blaauw}, and
  H.~{Kim}, ``An 879{GOPS} 243mw 80fps {VGA} fully visual cnn-slam processor
  for wide-range autonomous exploration,'' in \emph{2019 IEEE International
  Solid- State Circuits Conference - (ISSCC)}, 2019, pp. 134--136.

\bibitem{Ziyun-JSSC2019}
Z.~{Li}, J.~{Wang}, D.~{Sylvester}, D.~{Blaauw}, and H.~S. {Kim}, ``A 1920
  $\times$ 1080 25-{F}rames/s 2.4-{TOPS/W} low-power {6-D} vision processor for
  unified optical flow and stereo depth with semi-global matching,'' \emph{IEEE
  Journal of Solid-State Circuits}, vol.~54, no.~4, pp. 1048--1058, 2019.

\bibitem{PredictiveNet}
Y.~Lin, C.~Sakr, Y.~Kim, and N.~Shanbhag, ``{P}redictive{N}et: An
  energy-efficient convolutional neural network via zero prediction,'' in
  \emph{2017 IEEE International Symposium on Circuits and Systems (ISCAS)}, May
  2017, pp. 1--4.

\bibitem{RDSEC}
Y.~Lin, S.~Zhang, and N.~Shanbhag, ``{Variation-Tolerant Architectures for
  Convolutional Neural Networks in the Near Threshold Voltage Regime},'' in
  \emph{2016 IEEE International Workshop on Signal Processing Systems (SiPS)},
  Oct 2016, pp. 17--22.

\bibitem{liu2018adadeep}
S.~Liu, Y.~Lin, Z.~Zhou, K.~Nan, H.~Liu, and J.~Du, ``{On-demand deep model
  compression for mobile devices: A usage-driven model selection framework},''
  in \emph{Proceedings of the 16th Annual International Conference on Mobile
  Systems, Applications, and Services}.\hskip 1em plus 0.5em minus 0.4em\relax
  ACM, 2018, pp. 389--400.

\bibitem{network_slimming}
Z.~Liu, J.~Li, Z.~Shen, G.~Huang, S.~Yan, and C.~Zhang, ``Learning efficient
  convolutional networks through network slimming,'' in \emph{The IEEE
  International Conference on Computer Vision (ICCV)}, Oct 2017.

\bibitem{luo2017thinet}
J.-H. Luo, J.~Wu, and W.~Lin, ``Thinet: A filter level pruning method for deep
  neural network compression,'' in \emph{The IEEE International Conference on
  Computer Vision (ICCV)}, 2017, pp. 5058--5066.

\bibitem{thinet}
J.-H. Luo, J.~Wu, and W.~Lin, ``Thinet: A filter level pruning method for deep
  neural network compression,'' in \emph{The IEEE International Conference on
  Computer Vision (ICCV)}, 2017, pp. 5058--5066.

\bibitem{mao2017exploring}
H.~Mao, S.~Han, J.~Pool, W.~Li, X.~Liu, Y.~Wang, and W.~J. Dally, ``Exploring
  the granularity of sparsity in convolutional neural networks,'' in
  \emph{Proceedings of the IEEE Conference on Computer Vision and Pattern
  Recognition Workshops}, 2017, pp. 13--20.

\bibitem{nakkiran2015compressing}
P.~Nakkiran, R.~Alvarez, R.~Prabhavalkar, and C.~Parada, ``Compressing deep
  neural networks using a rank-constrained topology,'' 2015.

\bibitem{parashar2017scnn}
A.~Parashar, M.~Rhu, A.~Mukkara, A.~Puglielli, R.~Venkatesan, B.~Khailany,
  J.~Emer, S.~W. Keckler, and W.~J. Dally, ``Scnn: An accelerator for
  compressed-sparse convolutional neural networks,'' in \emph{2017 ACM/IEEE
  44th Annual International Symposium on Computer Architecture (ISCA)}.\hskip
  1em plus 0.5em minus 0.4em\relax IEEE, 2017, pp. 27--40.

\bibitem{qin2019accelerating}
Z.~Qin, D.~Zhu, X.~Zhu, X.~Chen, Y.~Shi, Y.~Gao, Z.~Lu, Q.~Shen, L.~Li, and
  H.~Pan, ``Accelerating deep neural networks by combining block-circulant
  matrices and low-precision weights,'' \emph{Electronics}, vol.~8, no.~1,
  p.~78, 2019.

\bibitem{sandler2018mobilenetv2}
M.~Sandler, A.~Howard, M.~Zhu, A.~Zhmoginov, and L.-C. Chen, ``Mobilenetv2:
  Inverted residuals and linear bottlenecks,'' in \emph{Proceedings of the IEEE
  Conference on Computer Vision and Pattern Recognition}, 2018, pp. 4510--4520.

\bibitem{PTPX}
Synopsys, ``{PrimeTime PX: Signoff Power Analysis},''
  \url{https://www.synopsys.com/support/training/signoff/primetimepx-fcd.html},
  accessed 2019-08-06.

\bibitem{tan2019efficientnet}
M.~Tan and Q.~V. Le, ``Efficientnet: Rethinking model scaling for convolutional
  neural networks,'' \emph{Proceedings of the 25th International Conference on
  Machine Learning}, 2019.

\bibitem{fp8}
N.~Wang, J.~Choi, D.~Brand, C.-Y. Chen, and K.~Gopalakrishnan, ``Training deep
  neural networks with 8-bit floating point numbers,'' in \emph{Advances in
  Neural Information Processing Systems 31}, S.~Bengio, H.~Wallach,
  H.~Larochelle, K.~Grauman, N.~Cesa-Bianchi, and R.~Garnett, Eds.\hskip 1em
  plus 0.5em minus 0.4em\relax Curran Associates, Inc., 2018, pp. 7675--7684.

\bibitem{wang2019e2}
Y.~Wang, Z.~Jiang, X.~Chen, P.~Xu, Y.~Zhao, Y.~Lin, and Z.~Wang,
  ``{E}2-{T}rain: Training state-of-the-art cnns with over 80\% energy
  savings,'' in \emph{Advances in Neural Information Processing Systems}, 2019,
  pp. 5139--5151.

\bibitem{wang2019dual}
Y.~Wang, J.~Shen, T.-K. Hu, P.~Xu, T.~Nguyen, R.~Baraniuk, Z.~Wang, and Y.~Lin,
  ``{Dual dynamic inference: Enabling more efficient, adaptive and controllable
  deep inference},'' \emph{IEEE Journal of Selected Topics in Signal
  Processing}, 2019.

\bibitem{wu2016quantized}
J.~Wu, C.~Leng, Y.~Wang, Q.~Hu, and J.~Cheng, ``Quantized convolutional neural
  networks for mobile devices,'' in \emph{Proceedings of the IEEE Conference on
  Computer Vision and Pattern Recognition (CVPR)}, 2016, pp. 4820--4828.

\bibitem{wu2018deep}
J.~Wu, Y.~Wang, Z.~Wu, Z.~Wang, A.~Veeraraghavan, and Y.~Lin, ``Deep $ k
  $-means: Re-training and parameter sharing with harder cluster assignments
  for compressing deep convolutions,'' \emph{Proceedings of the 25th
  International Conference on Machine Learning}, 2018.

\bibitem{Xu_2020}
\BIBentryALTinterwordspacing
P.~Xu, X.~Zhang, C.~Hao, Y.~Zhao, Y.~Zhang, Y.~Wang, C.~Li, Z.~Guan, D.~Chen,
  and Y.~Lin, ``{A}uto{DNN}chip: An automated dnn chip predictor and builder
  for both {FPGA}s and {ASIC}s,'' \emph{The 2020 ACM/SIGDA International
  Symposium on Field-Programmable Gate Arrays}, Feb 2020. [Online]. Available:
  \url{http://dx.doi.org/10.1145/3373087.3375306}
\BIBentrySTDinterwordspacing

\bibitem{yang2018dnn}
X.~Yang, M.~Gao, J.~Pu, A.~Nayak, Q.~Liu, S.~E. Bell, J.~O. Setter, K.~Cao,
  H.~Ha, C.~Kozyrakis \emph{et~al.}, ``Dnn dataflow choice is overrated,''
  \emph{arXiv preprint arXiv:1809.04070}, 2018.

\bibitem{WAGEUBN}
Y.~Yang, S.~Wu, L.~Deng, T.~Yan, Y.~Xie, and G.~Li, ``Training high-performance
  and large-scale deep neural networks with full 8-bit integers,'' 2019.

\bibitem{Yu_2017_CVPR}
X.~Yu, T.~Liu, X.~Wang, and D.~Tao, ``On compressing deep models by low rank
  and sparse decomposition,'' in \emph{The IEEE Conference on Computer Vision
  and Pattern Recognition (CVPR)}, July 2017.

\bibitem{yu2017compressing}
X.~Yu, T.~Liu, X.~Wang, and D.~Tao, ``On compressing deep models by low rank
  and sparse decomposition,'' in \emph{Proceedings of the IEEE Conference on
  Computer Vision and Pattern Recognition}, 2017, pp. 7370--7379.

\bibitem{zhang2016cambricon}
S.~Zhang, Z.~Du, L.~Zhang, H.~Lan, S.~Liu, L.~Li, Q.~Guo, T.~Chen, and Y.~Chen,
  ``Cambricon-x: An accelerator for sparse neural networks,'' in \emph{The 49th
  Annual IEEE/ACM International Symposium on Microarchitecture (MICRO)}.\hskip
  1em plus 0.5em minus 0.4em\relax IEEE Press, 2016, p.~20.

\bibitem{zhou2016dorefa}
S.~Zhou, Y.~Wu, Z.~Ni, X.~Zhou, H.~Wen, and Y.~Zou, ``Dorefa-net: Training low
  bitwidth convolutional neural networks with low bitwidth gradients,'' in
  \emph{Proceedings of the IEEE Conference on Computer Vision and Pattern
  Recognition}, 2018.

\bibitem{zhou2018cambricon}
X.~Zhou, Z.~Du, Q.~Guo, S.~Liu, C.~Liu, C.~Wang, X.~Zhou, L.~Li, T.~Chen, and
  Y.~Chen, ``Cambricon-s: Addressing irregularity in sparse neural networks
  through a cooperative software/hardware approach,'' in \emph{IEEE/ACM
  International Symposium on Microarchitecture (MICRO)}, 2018, pp. 15--28.

\end{thebibliography}


\end{document}